\documentclass[preprint,12pt,authoryear]{elsarticle}

\usepackage{graphicx,psfrag,color} 
  \graphicspath{ {./images/} }
\usepackage{amsmath}               
  \allowdisplaybreaks[1]           
\usepackage{amssymb}               
\usepackage{url}                   
\usepackage{rotating}              
\usepackage{multirow}              
\usepackage{lscape}                
\usepackage{tabularx}              
\usepackage{verbatim}              
\usepackage{footnote}              
\usepackage{float}                 
\usepackage{booktabs}              
\usepackage{lipsum}                
\usepackage[base]{babel}           
\usepackage{subcaption}            
\usepackage{siunitx}               
\usepackage{algorithm}             
\usepackage{algpseudocode}         
\usepackage{changepage}
\usepackage{hyperref}

\newcolumntype{P}[1]{>{\centering\arraybackslash\arraystretch}p{#1}}
\newcolumntype{M}[1]{>{\centering\arraybackslash}m{#1}}

\usepackage{amsthm}

\begin{document}

\begin{frontmatter}



\title{\texttt{naspaclust}: A hybridisation of spatial fuzzy clustering with nature-inspired optimisation}


 \author[inst1]{Bahrul Ilmi Nasution}
 \author[inst2,inst3]{Rezzy Eko Caraka}
 \author[inst4]{Robert Kurniawan}

 \affiliation[inst1]{organisation={Department of Social Statistics},
             addressline={The University of Manchester}, 
             city={Manchester},
             postcode={M13 9PL}, 
             state={Manchester},
             country={UK}}

 \affiliation[inst2]{organisation={Research Center for Data and Information Sciences, Research Organisation for Electronics and Informatics},Department and Organisation
             addressline={National Research and Innovation Agency}, 
             city={Bandung},
             postcode={40135}, 
             state={West Java},
             country={Indonesia}}

 \affiliation[inst3]{organisation={School of Economics and Business},Department and Organisation
             addressline={Telkom  University}, 
             city={Bandung},
             postcode={40257}, 
              state={West Java},
             country={Indonesia}}

 \affiliation[inst4]{organisation={Department of Statistical Computation},Department and Organisation
             addressline={Polytechnics Statistics STIS}, 
             city={Jakarta},
             postcode={13330}, 
              state={West Java},
             country={Indonesia}}

\begin{abstract}
We present an overview of R package called \texttt{naspaclust}, a tool for spatial fuzzy clustering improvement by hybridising the Fuzzy Geographically Weighted Clustering (FGWC) with nature-inspired optimisation algorithms. There are seven optimisation algorithms in general, namely Artificial Bee Colony, Flower Pollination Algorithm, Gravitational Search Algorithm, Harris-Hawk Optimisation, Intelligent Firefly Algorithm, Particle Swarm Optimisation, and Teaching-Learning Based optimisation. The optimisation algorithms available do not only use the basic algorithms, but also the modified version of several parts such as randomisation terms and updating rules. Furthermore, we also provide additional options by providing more distance metrics options based on the \texttt{ rdist} package. This study showed the implementation of the \texttt{naspaclust} package in various cases, as well as comparing the performance of the algorithms in the scope of basic algorithms. The \texttt{naspaclust} package allows to improve the performance of fuzzy clustering, mainly in spatial context.
\end{abstract}



\begin{keyword}
fuzzy geographically weighted clustering \sep geo-demographic analysis \sep nature-inspired optimisation \sep R package \sep spatial
\end{keyword}

\end{frontmatter}


\section{Introduction}
\label{sec:introduction}

    Clustering is a method used to divide a dataset into meaningful groups called clusters~\citep{caraka2020cluster}. Objects within the same cluster exhibit homogeneity while differing significantly from objects in other clusters. Clustering methods are frequently used for automatic classification in diverse fields, including medical sciences, engineering, and socio-economic analysis~\citep{han2011}. One notable socio-economic application of clustering is geo-demographic Analysis (GDA), a specialised area within Geographic Information Systems (GIS) that utilises spatial analysis techniques to identify and categorise demographic characteristics of specific regions~\citep{paez}. Such an analysis provides essential support for the development of targeted socio-economic policies~\citep{fgwcabc1,fgwcabc2}. The application of clustering techniques within GDA is particularly critical for policy making, as it allows efficient and precise assessments of regional characteristics, thus helping decision-makers who otherwise would spend considerable effort evaluating each region individually~\citep{Spiezia2007}.

    A widely used clustering method in GDA is fuzzy clustering, notably the Fuzzy C-Means (FCM) algorithm~\citep{kwon}. FCM is valued for its simplicity and improved performance over traditional hard clustering methods such as K-Means and Hard C-Means~\citep{singh2013}. Based on FCM, \cite{fgwc} introduced fuzzy geographic weighted clustering (FGWC), a spatially aware variant of FCM that incorporates geographical information into the clustering process, enabling geographically coherent clusters~\citep{fgwcabc2}. 

    However, despite its advantages, both FCM and FGWC are sensitive to the initial selection of cluster centroids, leading to suboptimal or unstable solutions~\citep{izakian,fgwcabc2}. This drawback requires enhancements to achieve globally optimal clustering results that better represent the underlying spatial-demographic structure of the data~\citep{fgwcabc2,liu2008,nayak}. Metaheuristic optimisation algorithms have emerged as promising solutions to overcome these initialisation issues by combining local and random search capabilities. To date, optimisation algorithms such as Ant Colony Optimisation (ACO)~\citep{fgwcaco,nurmala2015}, Gravitational Search Algorithm (GSA)~\citep{fgwcgsa}, Particle Swarm Optimisation (PSO)~\citep{fgwcpso}, and Firefly Algorithm (FA)~\citep{Nasution2020} have been integrated into FGWC with potential results.

    Moreover, previous studies indicate that modifying the distance metric used to calculate the membership of data points can significantly improve clustering outcomes~\citep{bezdek1984}. FCM and FGWC traditionally use Euclidean distance; however, evidence suggests that the use of alternative distance metrics could improve clustering performance~\citep{srivastava2011,singh2013,gueorguieva,wu2012}. Currently, the only publicly available package that specifically addresses FGWC is \texttt{spatialClust}, which provides limited options, primarily the classical FGWC and a version optimised with GSA~\citep{fgwcgsa}—and uses solely Euclidean distance. On the other hand, the \texttt{ppclust} package offers a variety of alternative distance metrics, but lacks integration with spatial clustering methods or optimisation algorithms.
    
    To address these gaps, this article aims to introduce our new R package called \texttt{naspaclust}, which comes from the abbreviation of nature-inspired spatial clustering. The main contribution of this study lies in the adaptation of several previous works on fuzzy clustering, particularly in the spatial context~\citep{Rao2012b, Rao2012, bhadoria2016, Wijayanto2015a,fgwcabc1,fgwcabc2,fgwcgsa,fgwcpso,nasution2018,Nasution2020,caraka2020cluster,nasutiontlbo,NASUTIONfpa,Sumargo2024}. Moreover, this package can behave simultaneously like its predecessors such as \texttt{spatialClust} regarding the optimisation in FGWC and \texttt{ppclust} regarding the distance adjustments.

    To address these gaps, this study introduces a novel package named \texttt{naspaclust} (nature-inspired spatial clustering), which significantly extends existing packages by incorporating multiple nature-inspired optimisation algorithms and diverse distance metrics into fuzzy spatial clustering methods. Specifically, \texttt{naspaclust} implements seven metaheuristic algorithms: Artificial Bee Colony (ABC), Gravitational Search Algorithm (GSA), Flower Pollination Algorithm (FPA), Firefly Algorithm (FA), Harris-Hawk Optimisation (HHO), Particle Swarm Optimisation (PSO), and Teaching-Learning-Based Optimisation (TLBO). In particular, recent advances such as intelligent search in FA~\citep{intfa,Nasution2020}, rapid progressive dives in HHO~\citep{Bairathi2019}, and elitism strategies in TLBO~\citep{Rao2012b, Rao2012, bhadoria2016,nasutiontlbo} are integrated to enhance algorithmic performance and computational efficiency. Furthermore, various randomisation strategies for algorithms like FPA and FA~\citep{randomfa,NASUTIONfpa} and numerous distance metrics from the \texttt{rdist} package~\citep{rdist}, including Minkowski and Canberra distances, have been implemented, improving cluster flexibility and quality based on specific data characteristics.
    
    Section~\ref{sec:back} describes the background information to provide an initial understanding of fuzzy clustering, FGWC, and the metaheuristic algorithms used in \texttt{naspaclust}, together with the research workflow. Section~\ref{sec:exp} explains the experimental setting used to compare different optimised FGWC algorithms on different data. Further, Section~\ref{sec:res} describes the simple description and usage of the \texttt{naspaclust} package, as well as the experimental results FGWC using standard optimisation algorithms. We provided an in-depth discussions of the package and the results in Section~\ref{sec:disc}. Last but not least, this study summarises everything in Section~\ref{sec:conc} and provides limitations and future directions of the implementation of~\texttt{naspaclust}.

\section{Background}
\label{sec:back}
\subsection{Fuzzy Clustering}
    In 1965, Lotfi Zadeh introduced a fuzzy logic concept that states that a certain subject has a true value between zero and one~\cite{ZADEH1965338}. In a clustering context, fuzzy logic works by defining that a subject can belong to more than one cluster with a specific calculated degree of membership~\citep{bezdek1981, han2011}. The most common fuzzy clustering method in GDA is Fuzzy C-means (FCM) which was introduced by~\cite{dunn} and developed by~\cite{bezdek1981}. FCM is a fuzzy clustering method that uses the degree of membership of an object with the object's distance to the center of the group~\citep{fgwc}. An object will be a member of a cluster if the object has the highest degree of membership in the cluster~\citep{nasution2018}. The objective of FCM is to minimise the weighted average of errors as a consequence of substituting data $x$ into the center of the cluster (centroid) $v$ using Equations~\eqref{eq:fcm_1} and~\eqref{eq:fcm_2}~\citep{bezdek1984}:

    \begin{equation}
        \label{eq:fcm_1}
        J_m(U,v) = \sum_{i=1}^{n} \sum_{j=1}^{c} \mu_{ij}^{m} d_{ij}^{2} 
    \end{equation}

    \begin{equation}
        \label{eq:fcm_2}
        d_{ij}^{2} = |x_i - v_h|_A^2 = (x_i - v_j)^T A(x_i - v_j)  
    \end{equation}

    With $m$ is an exponent weight called fuzzifier to determine the level of obscurity of a cluster, $\mu_{ij}$ is the membership value from the $i$-th observations and $j$-th cluster with constraint that sum membership value must be equal to 1 in each observation, $d_{ij}^2$ is the squared distance of the $i$-th observation to $j$-th centroid, $x_i$ is the $i$-th observation, $v_j$ is the $j$-th centroid, and $A$ is a $c \times c$ matrix. If $A$ is an identity matrix, the distance between the centroid and the data will satisfy the Euclidean distance formula. The minimisation of the objective function $J_m$ in the FCM algorithm is done by doing an iterative update of the centroid and the membership matrix using Equations~\eqref{eq:fcm_3} and~\eqref{eq:fcm_4}, respectively.

    \begin{equation}
        \label{eq:fcm_3}
        \mu_{ij} = \frac{1}{\sum_{k=1}^{c} (\frac{d_{ij}}{d_{ik}})^{\frac{2}{m-1}}} 
    \end{equation}

    \begin{equation}
        \label{eq:fcm_4}
        v_j = \frac{\sum_{i=1}^{n}(\mu_{ij})^m x_k}{\sum_{i=1}^{n}(\mu_{ij})^m} 
    \end{equation}

    This is done until the maximum difference value of $\mu_{ij}^{(t)}$ and $\mu_{ij}^{(t-1)}$ is smaller than the error.
    
\subsection{Fuzzy Geographically Weighted clustering}

   ~\cite{fengflowerdew} stated that the results of the fuzzy clustering are better if the neighborhood effect is added after the fuzzy clustering process. This is because of the membership values configuration after the fuzzy clustering process. However, this method still has disadvantages. First, the method ignores the influence of areas that do not have the same border. Second, the method shows a population influence that is of great concern in GDA~\citep{fgwc}. To overcome these weaknesses, \cite{fgwc} integrate these two processes in the FGWC by adding geographical influence in the form of distance between regions and populations within the region. This affects FGWC to affect the centroid value to make centroids spatially aware. The update of membership in the FGWC is calculated in each iteration using Equation~\eqref{eq:fgwc_1}~\citep{fgwc}.

    \begin{equation}
        \label{eq:fgwc_1}
       \mu'_{i}= \alpha \mu_{i} + \beta \frac{1}{A} \sum_{j}^{n} w_{ij} \mu_j
    \end{equation}

    Where $\mu'_{i}$ and $\mu_{i}$ is the modified and old members of the cluster in area $i$, respectively. The parameters $\alpha$ and $\beta$ are the scaling variables that influence the ratio between the original membership and the weighted (calculated) membership, respectively, with the relation $\alpha+\beta = 1$~\citep{fgwc}. The parameter $A$ is determined to ensure that the average weighted membership value is still in the range of $0$ and $1$. The $w_{ij}$ is the interaction weight of the area that is calculated based on spatial characteristics between areas using Equation~\eqref{eq:fgwc_2}~\citep{Nasution2020}.

    \begin{equation}
        \label{eq:fgwc_2}
        w_{ij}=\frac{(m_i m_j)^b}{d_{ij}^a}
    \end{equation}
    
    Where $m_i m_j$ are the total population in region $i$ and $j$, $d_{ij}$ is the distance between region $i$ and $j$, $a$ and $b$ represents the level of impact as a result of distance and population interaction between regions. These two parameters can be determined by the user. FGWC continues to configure the membership using the geographical effect for each iteration. 
    
    The FGWC can be represented into several algorithms based on the parameter configuration. When $\alpha=1$, it means that FGWC only uses the old membership as the configuration that is the same as FCM. On the other hand, when $\beta=1$, it means that the membership configuration is only considered by the spatial weight, which is the same as the neighborhood effects. Furthermore, the FGWC algorithm is explained in Algorithm~\ref{alg:fgwce}.

    \begin{algorithm}
    \caption{Fuzzy Geographically Weighted Clustering}
    \label{alg:fgwce}
    \hspace*{\algorithmicindent} \textbf{Input: } Data \boldmath$x$, spatial distance matrix \boldmath$d$, population matrix \boldmath$pop$, number of cluster $c$, fuzzifier $m$, ratio original membership $\alpha$, population effect $b$, spatial distance effect $a$, error tolerance $\varepsilon$ \\
    \hspace*{\algorithmicindent} \textbf{Output: }cluster of data, centroid \boldmath$v$, objective function $J_m (U,v;X)$ ,validation index $(PC,CE,SC,S,XB,IFV,Kwon)$
    \begin{algorithmic}
    \State $\beta=1-\alpha$
    \State Initialise Membership Matrix $\mu_0$
    \While {$max\mid\mu_{new}-\mu_{old}\mid\geq\varepsilon$}
    	\State calculate centroid using Equation~\eqref{eq:fcm_3}
    	\State calculate distance between data and centroid using Euclidean distance
    	\State calculate $\mu_{new}$ using Equation~\eqref{eq:fcm_4}
    	\State use geographical modification in Equation~\eqref{eq:fgwc_1} and Equation~\eqref{eq:fgwc_2} to update $\mu'$
    	\State calculate objective function using Equation~\eqref{eq:fcm_1} by setting $\mu_{ij}$ as $\mu'_{ij}$
    \EndWhile
    \State Calculate validation index of the cluster formation $(PC,CE,SC,S,XB,IFV,Kwon)$
    \State Obtained the cluster of each data using the best membership
    \end{algorithmic}
    \end{algorithm}

\subsection{Cluster Validation Measurement}
    Evaluating clustering results is essential to compare different models, select the optimal number of clusters, and ensure that the formed cluster reflects the underlying data structure\citep{bezdek1984}. In conventional clustering, indices such as the silhouette, Dunn, and Davies–Bouldin indices measure how compact and well separated the clusters are. However, these indices assume that each data point is assigned to one cluster and only one. These measurements is relevant when one wants to compare the fuzzy to non-fuzzy clustering methods.
    
    Due to the fuzzy nature, previous studies developed evaluation measurements that measure separation by involving the membership in each cluster. These measurements are relevant when selecting the appropriate fuzzy clustering methods in their case. This package provides seven well-known cluster validation measurements, namely: Partition coefficient (PC), Classification Entropy (CE), Partition Index (SC), Separation Index ($S$), Xie and Beni index ($XB$), IFV index, as well as Kwon index~\citep{Grekousis2012,toolbox,ifv,meteo}.

    \begin{itemize}
        \item \textit{Partition Coefficient (PC)}: Index that measures amount of overlapping between clusters. PC index value will always be between $\frac{1}{c}$ to 1. The optimal number of clusters is indicated by maximum PC value. PC index value is measured using Equation~\eqref{eq:pc}. 
        \begin{equation}
            \label{eq:pc}
            PC = \frac{1}{N}\sum_{i=1}^{c}\sum_{j=1}^{N}\mu_{ij}^{2}
        \end{equation}
        
        \item \textit{Classification Entropy (CE)}: CE represents the fuzziness between clusters. The value is measured using Equation~\eqref{eq:ce}~\citep{bezdek1984}. 
        \begin{equation}
            \label{eq:ce}
            CE = -\frac{1}{N}\sum_{i=1}^{c}\sum_{j=1}^{N}\mu_{ij} log_{a}(\mu_{ij})
        \end{equation}
        
        \item \textit{Partition Index (SC)}: SC is a ratio between the compactness to its cluster separation. A good cluster output can be seen from the minimum SC value~\citep{Grekousis2012}. SC Index can be calculated using Equation~\eqref{eq:sc}. 
        \begin{equation}
            \label{eq:sc}
            SC = \sum_{i=1}^{c} \frac{\sum_{j=1}^{N}(\mu_{ij})^m |x_j-v_i|^2}{N_i \sum_{k=1}^{c} |v_k-v_i|^2}
        \end{equation}
        
        \item \textit{Separation Index (S)}: S Index is a ratio between the objective function value to its minimum cluster separation. A good and optimal cluster partition is seen from the minimum S Index. S index value is calculated through Equation~\eqref{eq:s}~\citep{Grekousis2012}. 
        \begin{equation}
            \label{eq:s}
            S = \frac{\sum_{i=1}^{c}\sum_{j=1}^{N}(\mu_{ij})^2 |x_j-v_i|^2}{N min_{i,k} |v_k-v_i|^2}
        \end{equation}
        
        \item \textit{Xie and Beni’s Index (XB)}: XB is a ratio between the total variation within the cluster and the separation of clusters. A separated cluster is marked with aminimum XB value. The XB and SC index show how large the variation exists between clusters andhow clear they are separated. XB is measured using Equation~\eqref{eq:xb}~\citep{Grekousis2012}:
        \begin{equation}
            \label{eq:xb}
            XB = \frac{\sum_{j=1}^{c}\sum_{i=1}^{n} \mu_{ij}^m |x_i - v_j|^2}{n \min_{i,j} |x_i - v_j|^2}
        \end{equation}
        
        \item \textit{IFV Index}: IFV index is often used to validate fuzzy grouping with spatial data, because it is robust and stable~\citep{izakian,ifv}. A separated cluster is marked with maximum IFV value. IFV is measured using Equation~\eqref{eq:ifv}:
    
        \begin{equation}
            \label{eq:ifv}
            \begin{aligned}
            IFV = \frac{1}{c}\sum_{j=1}^{c}(\frac{1}{N}\sum_{k=1}^{N}\mu_{kj}^2[log_2 c-
            \frac{1}{N}\sum_{k=1}^{N}log_2 \mu_{kj}]^2)\frac{SD_{max}}{\overline{\sigma_D}}\\
            SD_{max} = max_{k \neq j}|v_k - v_j|^2\\
            \overline{\sigma_D} = \frac{1}{c} \sum_{j=1}^{c}(\frac{1}{N}\sum_{k=1}^{N}|x_k - v_j|^2)
            \end{aligned}
        \end{equation}

        \item \textit{Kwon Index}:~\cite{kwon} developed the $S$, $SC$, and $XB$ indices by adding tendencies which decreased monotonically when the number of clusters was very large and close to the amount of data. Better cluster partitions are marked with a minimum Kwon value. Kwon Index can be calculated using Equation~\eqref{eq:kwon}.

        \begin{equation}
            \label{eq:kwon}
            \begin{aligned}
            Kwon = \frac{\sum_{i=1}^{n}\sum_{j=1}^{c} \mu_{ij}^2 |x_i - v_j|^2}{\min_{j,k} |v_j - v_k|^2} +
            \frac{\frac{1}{c} \sum_{j=1}^{c} |v_j - \bar{v}|^2}{\min_{j,k} |v_j - v_k|^2}
            \end{aligned}
        \end{equation}
    \end{itemize}

    We selected these indices because they provide complementary perspectives on fuzzy clustering quality. In essence, they capture the degree of certainty of membership, the balance between intra‐cluster compactness and inter‐cluster separation, and the spatial characteristics of the clusters. These measurements ensure that our evaluation robustly reflects both the uncertainty inherent in fuzzy partitions and the underlying data structure.

\subsection{Nature-inspired optimisation features in \texttt{naspaclust}}
    There are many approaches to obtain the optimal parameter that produce an optimal objective function. The traditional approach uses numerical methods such as the Lagrange multiplier and Newton-Raphson~\citep{yang2014}. However, as the function becomes more complex, it is difficult to find the optimal solution using the traditional approach. Thus, an approach was developed by implementing the search algorithm based on the natural phenomenon in the world, known as nature-inspired optimisation algorithms~\citep{Fister2013}. This approach employs exploitation and exploration as the basic components to find the optimum solution~\citep{blum2003}. In the context of FGWC, we attempt to optimise the centroid to overcome the lack in FGWC. Thus, the position of $k$ -th swarms (e.g., particles, fireflies, flowers, etc.) in each algorithm represents the centroid with notation $v$. The hybridisation of FGWC and the nature-inspired optimisation algorithm can be seen in Algorithm~\ref{alg:fgwcopt}. The following Subsubsections explain briefly about the optimisation algorithms that are implemented in \texttt{naspaclust}.

    \begin{algorithm}
    \caption{Fuzzy Geographically Weighted Clustering with Nature-Inspired Optimisation Algorithm}
    \label{alg:fgwcopt}
        \hspace*{\algorithmicindent} \textbf{Input: } Data \boldmath$x$, spatial distance matrix \boldmath$d$, population matrix \boldmath$pop$, number of cluster $c$, fuzzifier $m$, ratio original membership $\alpha$, population effect $b$, spatial distance effect $a$, error tolerance $\varepsilon$, optimisation parameters. \\
        \hspace*{\algorithmicindent} \textbf{Output: }cluster of data, centroid \boldmath$v$, objective function $J_m (U,v;X)$ ,validation index $(PC,CE,SC,S,XB,IFV,Kwon)$
        \begin{algorithmic}
            \State Initialise swarm of centroids randomly $v_1,v_2,...,v_n$
            \State calculate objective function $f_1,\ f_2,\ldots,\ f_n$ using \eqref{eq:fcm_1}
            \State set $min(f_1,\ f_2,\ldots,\ f_n\ )$ as $f_{c_0}$ and $f_g$ and $min(v_1,v_2,...,v_n)$ as $v_{c_0}$ and $v_g$
            \State $t=0$
            \While {$f_{c_t}-f_{c_{t-1}}\geq\varepsilon$ or $t\le max_{iter}$}
                \State $t\gets t+1$
                \For {$k=1\rightarrow n$}
                    \State move each centroid to search based on the defined algorithm
                    \State calculate the membership of each swarm using Equation~\eqref{eq:fcm_3}
                    \State reconfigure the membership of each swarm spatially using Equation~\eqref{eq:fgwc_1} and Equation~\eqref{eq:fgwc_2}
                    \State update the centroid of each swarm using Equation~\eqref{eq:fcm_4}
                    \State calculate fitness of each swarm using Equation~\eqref{eq:fcm_1}
                \EndFor
                \State update optimisation parameters
                \State set $min(f_1,\ f_2,\ldots,\ f_n\ ) as f_{c_t}$ and $max(v_1,v_2,...,v_n)$ as $v_{c_t}$
                \If {$f_g\ <f_{c_t}$} 
                    \State update $f_g$ and $v_g$
                \EndIf
            \EndWhile
            obtain the optimum centroid $v_g$ with the objective value $f_g$
        \end{algorithmic}
    \end{algorithm}

\subsubsection{Artificial Bee Colony}
    ABC algorithm is inspired from the bee colony behavior. The bee colony consists of three groups: worker, onlooker, and scout bees~\citep{Kuo2018}. A worker bee only places a position of food source which is represented by the value of objective function. The movement relies on random bees around and is represented in \eqref{eq:abc1}~\citep{yang2014}. In \texttt{naspaclust}, we add the global exploration term using the global best $gbest$ inspired by the PSO algorithm~\citep{Shahrudin2014} for movement as seen in Equation~\eqref{eq:abc1}.
        \begin{equation}
            \label{eq:abc1}
            v_k^{t+1}=v_k^t+\phi\left(v_k^t-v_h^t\right)+\psi(gbest^t-v_k^t),\ k\neq h	    
        \end{equation}
    
    $v$ represents the food position of the bee, i.e. the centroid. $k$ and $h$ represent the determined index and random index of the chosen bee. $\phi$ is the control factors for the fabrication of foods that is randomly generated from [-1, 1]~\citep{Karaboga2007}. $psi$ is the control parameter for the PSO term, which is random from [0,1.5]. It holds when \texttt{pso=TRUE}. Otherwise, $\psi$ will be set to 0 and the bees move using basic movement. The position of the food source is updated if the food source is preferable to the previous source~\citep{Karaboga2011}. On the other hand, the onlooker bees consider the odds to select the food sources. The odds are directly proportional to the objective function and are represented in Equation~\eqref{eq:abc2}. 
        \begin{equation}
            \label{eq:abc2}
            p_k=\frac{fitness_k}{\sum_{h=1}^{n}{fitness_h}}	 	    
        \end{equation}
    
    Subsequently, the onlooker bees are selected by sorting the worker bees based on the probability available and updating them again using \eqref{eq:abc1}. If there is a $k$-th bee that cannot improve the food source to a certain extent, a new bee will come to replace the old one using Equation~\eqref{eq:abc3}.
        \begin{equation}
            \label{eq:abc3}
            v_k^{t+1}=v_{min}^t+\epsilon(v_{max}^t-v_{min}^t)	    
        \end{equation}
    
    Where $v_{min}^t$ and $v_{max}^t$ are the centroid with minimum and maximum food positions, respectively. Meanwhile, $\epsilon$ is a random walk between [0, 1]. This process is repeated until the termination criteria are found. 

\subsubsection{Flower Pollination Algorithm}
    FPA has two pollination processes in motion: cross-pollination and self-pollination~\citep{Yang2012}. Cross pollination is defined as a global pollination performed by pollinators who can fly to the $k$-th flower in long distances represented by Levi flights~\citep{carakafpa}. Mathematically speaking, this motion is calculated in Equation~\eqref{eq:fpa1}.
        \begin{equation}
            \label{eq:fpa1}
            v_k^{t+1}=v_k^t+\gamma L(\lambda)(g_{best}-v_k^t)	
        \end{equation}
    
    With $\gamma$ being the insect movement scaling factor and $L\left(\lambda\right)$ is a Levy distributed random number with scalar factor $\lambda$~\citep{NASUTIONfpa}. On the other hand, local or self-pollination is the pollination of a certain flower that occurred when there is no pollinators~\citep{yang2014}. The local pollination is represented in Equation~\eqref{eq:fpa2}
        \begin{equation}
            \label{eq:fpa2}
            v_k^{t+1}=v_k^t+\epsilon\left(v_h^t-v_k^t\right),\ h\neq k
        \end{equation}
    
    $\epsilon$ is the random pollination factor that is basically uniform [0,1]. We add the flexibility of the random pollination factor in \texttt{naspaclust} by adding more distribution options inspired by randomised FA. Those distributions become uniform, normal, Levy flights, and two types of chaotic maps (logistic and kent) (see~\cite{randomfa} for full details). Moreover, chaotic randomisation has also been implemented in FPA, particularly in a clustering context~\citep{AroraAnand2017,Kaur2020, NASUTIONfpa}. To determine whether a flower receives global or local pollination, a random number between 0 and 1 is compared to the probability of switch $p$. If the random number is less than $p$, the flower receives global pollination. Otherwise, the flower does local pollination~\citep{Dhal2019}.

\subsubsection{Gravitational Search Algorithm}

    Gravitational Search Algorithm (GSA) is an algorithm found by~\cite{rashedi2009}. This algorithm is inspired by a very well-known law of physics, Newton's law of gravity. This law states that some objects attract each other with gravitational force, and that gravitational force makes them move towards objects with heavier masses. Mathematically speaking, the formula is represented in Equation~\eqref{eq:newton_law}.
    \begin{equation}
        \label{eq:newton_law}
        F_{12}=G\frac{M_{1}\times M_{2}}{R^2},
        a_{1}=F_{12}/M_{1}
    \end{equation}
    
    with $M_{1}$ and $M_{2}$ representing the gravitational mass of particles 1 and 2. GSA starts from the initialisation of the agents (Masses) with the position $k$ -th agent in the $d$ number of dimensions defined in Equation~\eqref{eq:position}.
    \begin{equation}
        \label{eq:position}
        v_k=v_k^1,v_k^2,...,v_k^d, k=1,2,...N,
    \end{equation}
    
    at a certain time $t$, the force acting on the mass $k$ to the mass $h$ in each dimension $d$ is a modification of formula \eqref{eq:mass_force} 
    \begin{equation}
        \label{eq:mass_force}
        F_{kh}^d(t)=G(t)\frac{M_{k}(t)\times M_{h}(t)}{R_{kh}(t)+\varepsilon}(v_k^d(t)-v_h^d(t)),
    \end{equation}
    
    with $G(t)$ is the gravitational constant at time $t$, $M_k$ and $M_h$ is the gravitational mass of agent $k$ and $h$, respectively. $R_{kh}(t)$ is the Euclidean distance between agent $k$ and $h$, and $\varepsilon$ is a small constant. The individual mass of $k$-th agent is evaluated using the difference between the fitness of $k$-th agent and the worst fitness of the population compared to the difference between the best and worst fitness of the population. The gravitational mass of each agent is calculated using its ratio of its inertial mass $m_k$ compared to the sum of all individual masses in the population~\citep{oliveira2018}. Mathematically speaking, the formula for the update of the masses is written in Equation~\eqref{eq:mass}.
    \begin{equation}
        \label{eq:mass}
        \begin{aligned}
            m_k=\frac{fit_k(t)-worst(t)}{best(t)-worst(t)}\\
            M_k=\frac{m_k}{\sum_{j=1}^{N} m_h(t)}
        \end{aligned}
    \end{equation}
    
    The acceleration of agent $k$ at time $t$ in a certain direction $a_k^d(t)$ can be obtained from the force that acts on agent $k$ in a certain dimension $F_k^d(t)$. It is a randomly weighted sum of each dimension of the forces undertaken by the other agents. The acceleration and force formula are given in Equation~\eqref{eq:accel}.
    \begin{equation}
        \label{eq:accel}
        \begin{aligned}
            F_k^d(t)=\sum_{j=1,j\neq i}^{N} {rand}_hF_{kh}^d(t)\\
            a_k^d(t)=F_k^d(t)/M_{ii}(t)
        \end{aligned}
    \end{equation}
    
    Moreover, the next velocity of the agent $v_k^d(t+1)$ can be obtained from the sum of the current velocity $v_k^d(t)$ and its acceleration at time $t$. Finally, the velocity and position can be updated in Equation~\eqref{eq:velocity}.
    \begin{equation}
        \label{eq:velocity}
        \begin{aligned}
            \omega_k^d(t+1)={rand}_k \times \omega_k^d(t)+a_k^d(t)\\
            v_k^d(t+1)=v_k^d(t)+\omega_k^d(t)
        \end{aligned}
    \end{equation}
    
    This package adds other options for using a new technique of GSA or not. This new technique is done after the first position update~\citep{li2017gsa}. The update formula can be seen in Equation~\ref{eq:gsa_new}.
    
    \begin{equation}
        \label{eq:gsa_new}
        v_k^d(t+1)=\upsilon_k+\varsigma_k\times Z_k
    \end{equation}
    
    with $\upsilon$ and $\varsigma$ is the mean and standard deviation between the current agent and its best position personally and globally in the current iteration. Meanwhile, $Z$ is the randomisation constant for the agent population. This new technique can balance exploration and exploitation as well as maintain the heterogeneity between agents. The formula for $\upsilon$, $\varsigma$, and $Z$ can be seen in Equation~\eqref{eq:gsa_new2}.
    \begin{equation}
        \label{eq:gsa_new2}
        \begin{aligned}
            \upsilon_k = \frac{v_k^d(t)+pbest_k^d(t)+gbest^d(t)}{3}\\
            \varsigma_k = \sqrt{\frac{(v_k^d(t)-\upsilon)^2+(pbest_k^d(t)\upsilon)^2+(gbest^d(t)\upsilon)^2}{3}}\\
            Z_k = (-2 \ln{c_1})^{0.5}\times \cos{(2\pi c_2)}
        \end{aligned}
    \end{equation}

\subsubsection{Harris-Hawk Optimisation}
    HHO is an optimisation algorithm created by~\cite{Bairathi2019}  which was then developed by~\cite{Heidari2019} by increasing the exploitation ability of each eagle to find the optimum solution known as a rabbit~\citep{setiawanrobert}. Basically, this algorithm has 2 stages, namely exploration and exploitation. Before this stage, the energy of each eagle will be evaluated to determine whether to enter the exploration or evaluation stage~\citep{Singh2020a} using Equation~\eqref{eq:hhoexplore}.
        \begin{equation}
            \label{eq:hhoexplore}
             E_k=2E_{0,k}(1-\frac{t}{T})   
        \end{equation}
    
    With $E_{0,k}$ is the initial energy of $k$-th Harris-Hawk eagle to hunt the rabbit. Based on the equation, if $\left|E_k\right|>1$, the eagle still has more energy, thus it explores different regions to hunt the rabbit based on the other family members when $q<0.5$ or the random tall tree when $q\geq0.5$. The exploration formula is described in Equation~\eqref{eq:hhoexplore2}.
    \begin{equation}
        \label{eq:hhoexplore2}
        v_k^{t+1}=
        \begin{cases}
            v_h^t-r_1\left|v_h^t-2r_2v_k^t\right| & q\geq0.5\\
            v_{rabbit}^t-v_m^t-r_3\left(lb+r_4\left(ub-lb\right)\right) & q<0.5
        \end{cases}
    \end{equation}
    
    where $v_k^t$ and $v_k^{t+1}$ is the eagle that will do the exploration in iteration $t$ and $t + 1$, respectively. $v_h^t$ is a randomly chosen eagle position among the current population. $r_1,r_2,r_3,r_4$ and q is a random number between [0, 1]. Then $v_{rabbit}^t$ is the position of the rabbit, i.e, the best centroid at iteration $t$, and $v_m^t$ is the average position of all eagles. $lb$ and $ub$ is the lower and upper bound of the parameters. 
    
    When $\left|E_k\right|<1$, the eagle starts to run out of energy and enters the exploitation stage~\citep{Bairathi2019}. At this stage, the energy of each eagle is reevaluated by a limit of 0.5 and generating a random number. If $\left|E_k\right|\geq0.5$, the eagle's energy is still sufficient to attack the prey slowly (soft Besiege). Otherwise, the eagle will take a surprise attack on the rabbit (hard Besiege)~\citep{Heidari2019}. Meanwhile, the value of $r$ is generated to determine whether the eagle has the additional ability to select an attack pattern to get the best prey known as progressive rapid dives~\citep{Singh2020a,Sumargo2024}. If $r\geq0.5$, the exploitation stage of the Harris-hawk eagle is calculated by Equation~\eqref{eq:hhoexploit1}~\citep{Heidari2019}.

        \begin{equation}
            \label{eq:hhoexploit1}
            v_k^{t+1}=
            \begin{cases}
            v_{rabbit}^t-v_k^t-E_k|2(1-r_5)v_{rabbit}^t-v_k^t| & \left|E_k\right|\geq0.5\\
            v_{rabbit}^t-E_k|v_{rabbit}^t-v_k^t| & \left|E_k\right|<0.5
            \end{cases}
        \end{equation}
    Meanwhile, when $r<0.5$, the exploitation formula is calculated in \eqref{eq:hhoexploit2}
        \begin{equation}
            \label{eq:hhoexploit2}
            v_k^{t+1}=
            \begin{cases}
            Y & if\ fitness\left(Y\right)<fitness_k\\
            Z & i f\ fitness\left(Z\right)<fitness_k
            \end{cases}
        \end{equation}
    With
        \begin{equation}
            \label{eq:hhoexploit3}
            \begin{aligned}
            Y=
            \begin{cases}
                v_{rabbit}^t-E_k|2(1-r_5)v_{rabbit}^t-v_k^t| & \left|E_k\right|\geq0.5 \\
                v_{rabbit}^t-E_k|2(1-r_5)v_{rabbit}^t-v_m^t| & \left|E_i\right|<0.5
            \end{cases} \\
            Z=Y+S\times LF
            \end{aligned}
        \end{equation}
    Where $S$ is a random number as many as dimensions and $LF$ is the generated number of Levy flights function. 

\subsubsection{Intelligent Firefly Algorithm}
    The other algorithm is named IFA. It is an FA algorithm modified by~\citep{intfa} by adding a firefly sorting system based on the best solution. Prior to this algorithm, the movement of existing fireflies is dependent on the selected fireflies, which accelerates the computation and increases the ability to achieve the global optimum~\citep{Nasution2020}. Consequently, an additional parameter $\phi$ is needed, which reflects the proportion of the best fireflies selected. However, this study uses the number of intelligent fireflies $n_{intel}$, which can be determined easier and avoid rounding when using proportion. The movement of a new firefly is updated by Equation~\eqref{eq:ifa1}.
        \begin{equation}
            \label{eq:ifa1}
            v_k^{t+1}=v_k^t+\beta_0e^{-\gamma r_{kh}^2}\left(v_{intel,h}^t-v_k^t\right)+\alpha^t\epsilon_k^t
        \end{equation}
     
     $v_k^{t+1}$ and $v_k^t$ is the positions of the $k$-th firefly at times $t+1$ and $t$, respectively. The second term of the equation represents the attraction that is affected by the attractiveness ($\beta$), light absorption ($\gamma$), and the distance between 2 fireflies ($r_{kh}$). In the second term, there is also $v_{intel,h}^t$ which represents the h-th intelligent firefly~\citep{intfa}. Next, the third term is called randomisation, with $\alpha$ being the randomisation constant and $\epsilon_k^t$ the random movement of $k$-th firefly, which by default is normally distributed~\citep{yang2014}. This package adds flexibility to IFA to choose the types of update $\alpha$ which are summarised from previous studies~\citep{yang2009,yang2014,intfa} (see Table~\ref{tab:alphaifa} for details). Moreover, parallel to FPA, there are five types of random walk in \texttt{naspaclust}~\citep{yang2014, randomfa}.
    
    \begin{table}[htbp]
        \centering
        \caption{Randomisation constant update formula}
        \begin{tabular}{M{0.2\linewidth}M{0.3\linewidth}}
        \toprule
        \textbf{Update type} & \textbf{Formula} \\ \midrule
        1 & $10^{-5}+(\alpha-10^{-5})\times e^{t}$ \\
        2 & $\alpha\times rand^{t}$ \\
        3 & $1-(\frac{10^{-4}}{9})^{\frac{1}{max_{iter}}}$ \\
        4 & $\left(1,11\times{10}^{-4}\right)^\frac{5}{max_{iter}}{\times\alpha}$\\
        5 & $\alpha\times \frac{t}{max_{iter}}$ \\
        \bottomrule
        \end{tabular}
        \label{tab:alphaifa}
    \end{table}

\subsubsection{Particle Swarm Optimisation}
    PSO is a simple algorithm. Since the movement of each particle depends on the velocity of the particle~\citep{Tsekouras2013}, each particle would remember their personal best position $p_{k,best}$ and also learn from the best position among all particles at the current iteration $g_{best}$~\citep{Kennedy}. The update position of each particle can be calculated simply in Equation~\eqref{eq:pso1}.
        \begin{equation}
            \label{eq:pso1}
            \begin{aligned}
            \omega_k^{t+1}=\theta^t \omega_k^t+c_1\epsilon_1\left(g_{best}-v_k^t\right)+c_2\epsilon_2\left(p_{k,best}-v_k^t\right)\\
            v_k^{t+1}=v_h^t+\omega^{t+1}
            \end{aligned}
        \end{equation}
    
    $\omega^{t+1}$ and $\omega^t$ represent the velocity at the $t+1$ and $t$-th iteration, respectively. The velocity will be adjusted according to the maximum velocity $\omega_{max}$ if it exceeds the maximum limit~\citep{caraka2019pso}. $\theta$ is the inertial weight, then $c_1$ and $c_2$ represent the global and personal exploration acceleration constant. $\epsilon_1$ and $\epsilon_2$ are a random number with [0, 1] indicating the 'craziness' of the particle~\citep{Kennedy}. Meanwhile, $v^{t+1}$ and $v^t$ show the particle's position at the $t+1$ and $t$-th iteration. This package added some options to update the inertial weight because this method gives the smallest error among the various weights in optimising PSO~\citep{Bansal2011}. The update can be seen in Table~\ref{tab:inertpso}.

    \begin{table}[htbp]
        \centering
        \caption{Inertia Weight Strategies}
        \begin{tabular}{M{0.3\linewidth}M{0.5\linewidth}}
        \toprule
        \textbf{Inertia Weight} & \textbf{Formula}  \\ \midrule
        Constant Inertia Weight & $w = w_{max}$ \\
        Chaotic Inertia Weight & $w=(w_{max}-w_{min})\times (1-\frac{iter}{max_{iter}})+w_{min}*z$ \\
        Simulated Annealing Inertia Weight & $w_{min}+(w_{max}-w_{min})\times 0.95^{iter-1}$ \\
        Natural Exponent 1 Inertia Weight & $w_{min}+(w_{max}-w_{min})\times exp(-\frac{iter}{\frac{max_{iter}}{10}})$\\
        Natural Exponent 2 Inertia Weight & $w_{min}+(w_{max}-w_{min})\times exp(-\frac{iter}{\frac{max_{iter}}{4}}^2)$\\
        \bottomrule
        \end{tabular}
        \label{tab:inertpso}
    \end{table}

\subsubsection{Teaching-Learning Based Optimisation}
    TLBO is an algorithm that is inspired by the classroom system in education: teaching and learning~\citep{nasutiontlbo}. Generally, this algorithm consists of a number of students with learnt lessons represented by objective functions. This algorithm has 2 phases of movement~\citep{nayak}. The first phase is teacher selection, where the teacher is the smartest student in the class and will teach other students~\citep{Rao2012}. The ability of the students to absorb the lessons depends on the gap between the teacher and the average student $M$ known as $DM$. In the following, the gap is used to update the student's knowledge~\citep{Rao2012b}. Mathematically, the teaching phase is explained in Equation~\eqref{eq:tlbo1}~\citep{Rao2012}.
        \begin{equation}
            \label{eq:tlbo1}
            \begin{aligned}
            DM_k=r_k(v_{best}-{Tf}_kM)\\
            {Tf}_k=round[1+r]\\
            v_k^{\prime t}=v_k^t+DM_k
            \end{aligned}
        \end{equation}
    
    With $v_{best}$ is the best centroid, i.e. teacher among the population. ${Tf}_k$ is the random teaching factor that represents the learning capacity between students, $r$ is a random number generated from [0,1]. $v_k^t$ and $v_k^{\prime t}$ are the current and next positions. After the teacher shares its knowledge to the students, the second phase is the improvement of the knowledge by the students from the interaction with random fellow students. Smarter students will teach students below them. In the case of minimising the objective function, the updating of the position of each student is mathematically determined in Equation~\eqref{eq:tlbo2}.
        \begin{equation}
            \label{eq:tlbo2}
            Y=
            \begin{cases}
            v_k^{\prime t}+r_k\left(v_k^{\prime t}-v_h^{\prime t}\right) & fitness_k<fitness_h\\
            v_k^{\prime t}+r_k\left(v_h^{\prime t}-v_k^{\prime t}\right) & fitness_k>fitness_h
            \end{cases}
        \end{equation}
    $v_k^{t+1}=v_k^{\prime\prime t}$ if the current objective function is better than the teaching phase. Before continuing to the next iteration, we can decide whether to use the elitism steps or not. The elitism algorithm works by replacing a certain number of bad students with the best ones and modifying duplicated students to maintain diversity~\citep{Rao2012b,nasutiontlbo}.

\section{Experimental Setup}
\label{sec:exp}
To make an easier introduction of the R package, this study attempts to implement the \texttt{naspaclust} package into several data benchmarks that the analyses was implemented using FGWC as follows:

\begin{itemize}
    \item \textit{Case 1: 2010 Indonesia population census}~\citep{fgwcabc2}: This data contains 110 characteristics which represents the demographic condition in Indonesia at provincial level. The data was obtained from 2010 Indonesia population census. This data represents the condition where the number of regions is small with high dimensionality.
    \item \textit{Case 2: East Java dengue fever}~\citep{putra2017}: This data consists of 9 dengue fever indicators in East Java, Indonesia at district level. The variables in this study includes the percentage of dengue fever cases, bamboo walls, poor households, unhealthy homes, non-healthy households, larva-free houses, rainwater storage, number of flood events, and houses with palm fiber roofs. This data represents the condition where the number of regions is slightly larger than the variables.
    \item \textit{Case 3: Indonesia social vulnerability}~\citep{Nasution2020,KURNIAWANsvdata}: This data consists of 16 characteristics of social vulnerability of Indonesia in district level, namely the socioeconomic, demography, and the housing conditions. This data represents the condition where the number of regions is much larger than the dimensions.
\end{itemize}

We compared each algorithms performance by using the three datasets mentioned before. Despite the objective function, we also compare the validation indices using each combination. Considering the stochastic nature of metaheuristic optimisation, we run the simulation 100 times for each combination of dataset, algorithm, and number of clusters, which are from 2 to 6. After running the algorithms, we used the following evaluations:

\begin{itemize}
    \item \textit{Success final cluster output}: We evaluated the final cluster output to assess whether the algorithms produced the expected results. We consider the "success" term when the algorithm can produce the expected number of clusters as defined at the beginning. For example, if one wants to form four clusters, then the algorithm should be able to group the data into four clusters instead of three or even less. We used the number of successes obtained for each algorithm in every number of clusters.
    \item \textit{Quantitative evaluation}: we evaluate the performance based on the average performance measurement for the appropriate results in each algorithm. Note that in this evaluation, we only consider values that succeed in producing the expected number of clusters. When an algorithm did not produce an appropriate number of clusters, the measurement values in the corresponding number of clusters will be denoted by "NA" to make easier display.
\end{itemize}

With all the evaluation criteria, and considering the different data characteristics, we can compare the advantages and disadvantages of each algorithm in optimising FGWC, especially given the number of data.

In this study, similar FGWC parameters were used: $2\leq c \leq6$, $m=2$, $a=b=1$, and $\epsilon=10^{-6}$. On the other hand, this study employs 10 populations for each optimisation algorithm with 10 consecutive indifferent changes and 1000 iterations as the termination criteria. The detailed optimisation algorithms parameter used in this study are explained in Table~\ref{tab:optpar}.

\begin{center}
\begin{table}[t]%
\centering
\caption{Optimisation algorithms parameter used in the study}%
\begin{tabular}{M{0.3\linewidth}M{0.4\linewidth}}
\toprule
\textbf{Algorithm} & \textbf{Parameters} \\
\midrule
Artificial Bee Colony & $n_{onlooker}=5, limit=10$  \\
Flower Pollination Algorithm & $p=0.8,  \gamma=1.2, \lambda=0.1$  \\
Gravitational Search Algorithm & $G_0=2, v_{max}=0.7$  \\
Harris-Hawk Optimisation & $q=0.5$\\
Intelligent Firefly Algorithm & $n_{intel}=2, \gamma_f=\beta_0=1$, update type = 4\\
Particle Swarm Optimisation & $v_{max}=0.7, c_1=0.7, c_2=0.6, w_{max}=0.9, w_{min}=0.2, z=0.3$, weight update  = chaotic inertia weight\\
Teaching-Learning Based Optimisation & - \\
\bottomrule
\end{tabular}
\label{tab:optpar}
\end{table}
\end{center}

\section{Results} 
\label{sec:res}
\subsection{A brief overview of \texttt{naspaclust}}

This package contains the function to perform clustering using FGWC. Furthermore, we add several improvements to address the disadvantages in fuzzy clustering by using a nature-inspired optimisation algorithm. (fgwc and optimisation) This package also provides data sets from the 2010 Indonesian Population Census which are aggregated into 33 provinces and contain 110 characteristics. This data has been implemented by~\citep{fgwcabc2} in evaluating the performance of FGWC using the optimisation algorithm. The detail can be seen in~\citep{Wijayanto2015a}. 

In this section, we attempted to provide an explanation of how to use \texttt{naspaclust} package. First of all, we need to import the library. The \texttt{data} part is optional. If one wants to use their own data, they should have three components, namely main data, distance information, and population.

\begin{verbatim}
install.packages("naspaclust")
library(naspaclust)
data("census2010") ## main data
data("census2010dist") ## distance information
data("census2010pop") ## population
\end{verbatim}

After calling the data, we can define the parameters of the FGWC as well as the optimisation algorithm if we use the optimisation algorithm. This example shows a simple implementation of FGWC using the ABC optimisation algorithm. We get several return values related to the FGWC once the run has ended (see~\cite{naspaclustpkg} for details).

\begin{verbatim}
param_fgwc <- c(kind="v",ncluster=3,m=2,distance="minkowski",
                order=3,alpha=0.5,a=1.2,b=1.2,
                max.iter=1000,error=1e-6,randomN=10)
abc_param <- c(vi.dist="normal",npar=5,pso=FALSE,same=15,n.onlooker=5,
               limit=10) 
res <- fgwc(census2010,census2010pop,census2010dist,"abc",
            param_fgwc,abc_param) 
\end{verbatim}

\subsection{Can FGWC produce the expected number of clusters?}
\label{ss:clustfor}

\begin{figure}[htbp]
    \centering
    \begin{subfigure}[b]{0.48\textwidth}
        \centering
        \includegraphics[width=\textwidth]{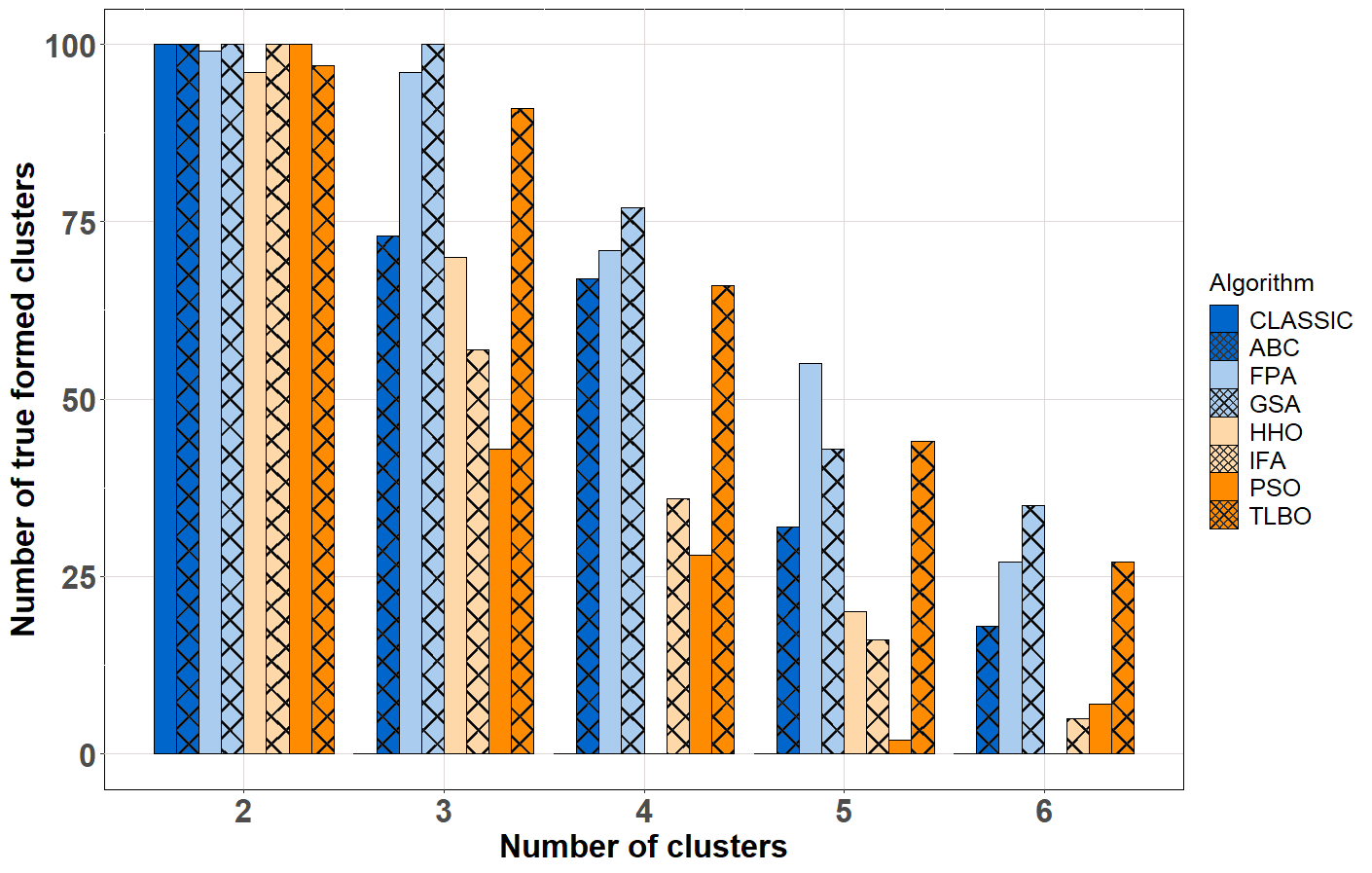}
        \caption{Case 1}
        \label{subfig:eval_1}
    \end{subfigure}
    \hfill
    \begin{subfigure}[b]{0.48\textwidth}
        \centering
        \includegraphics[width=\textwidth]{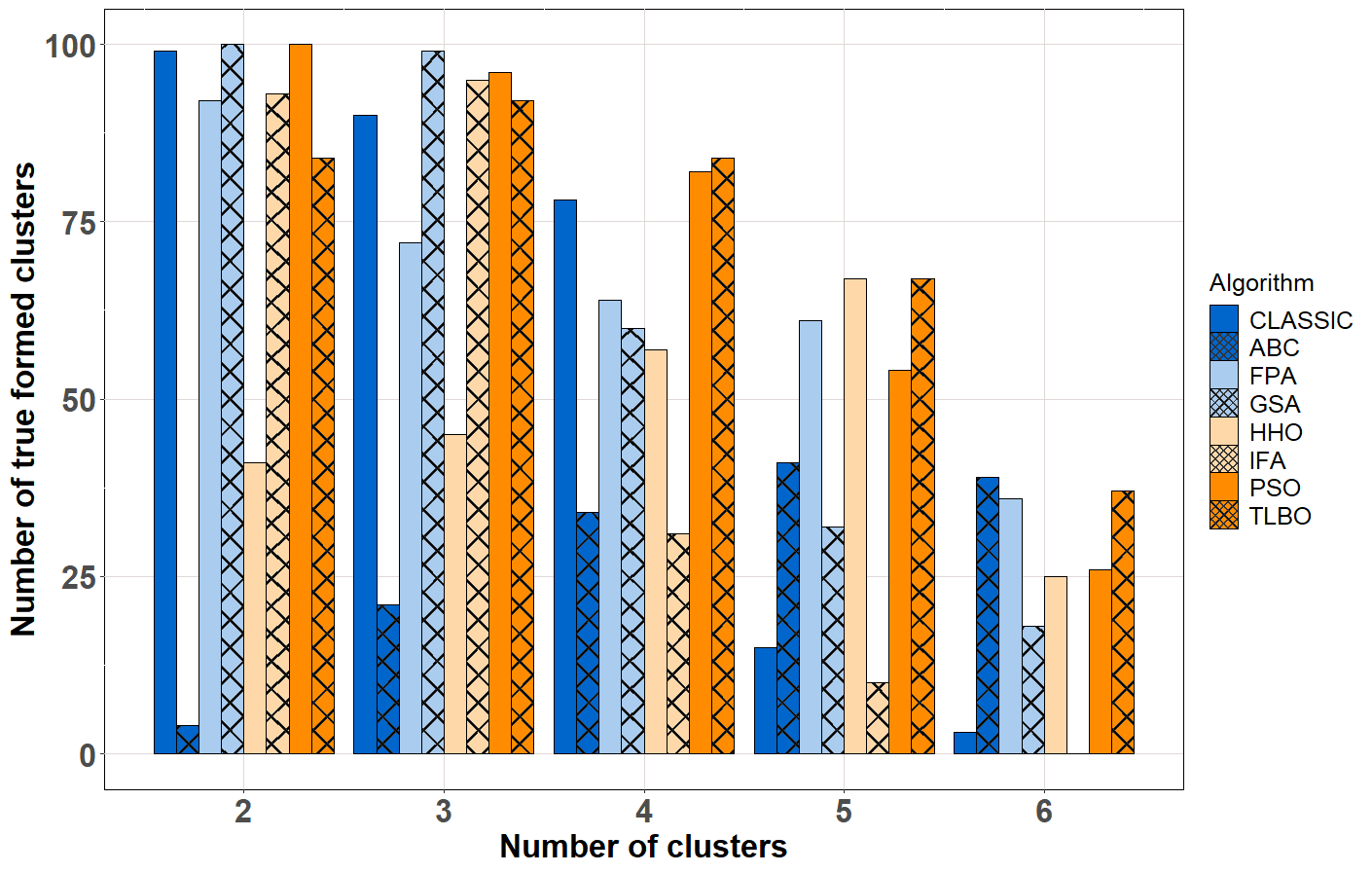}
        \caption{Case 2}
        \label{subfig:eval_2}
    \end{subfigure}
    \hfill
    \begin{subfigure}[b]{0.45\textwidth}
        \centering
        \includegraphics[width=\textwidth]{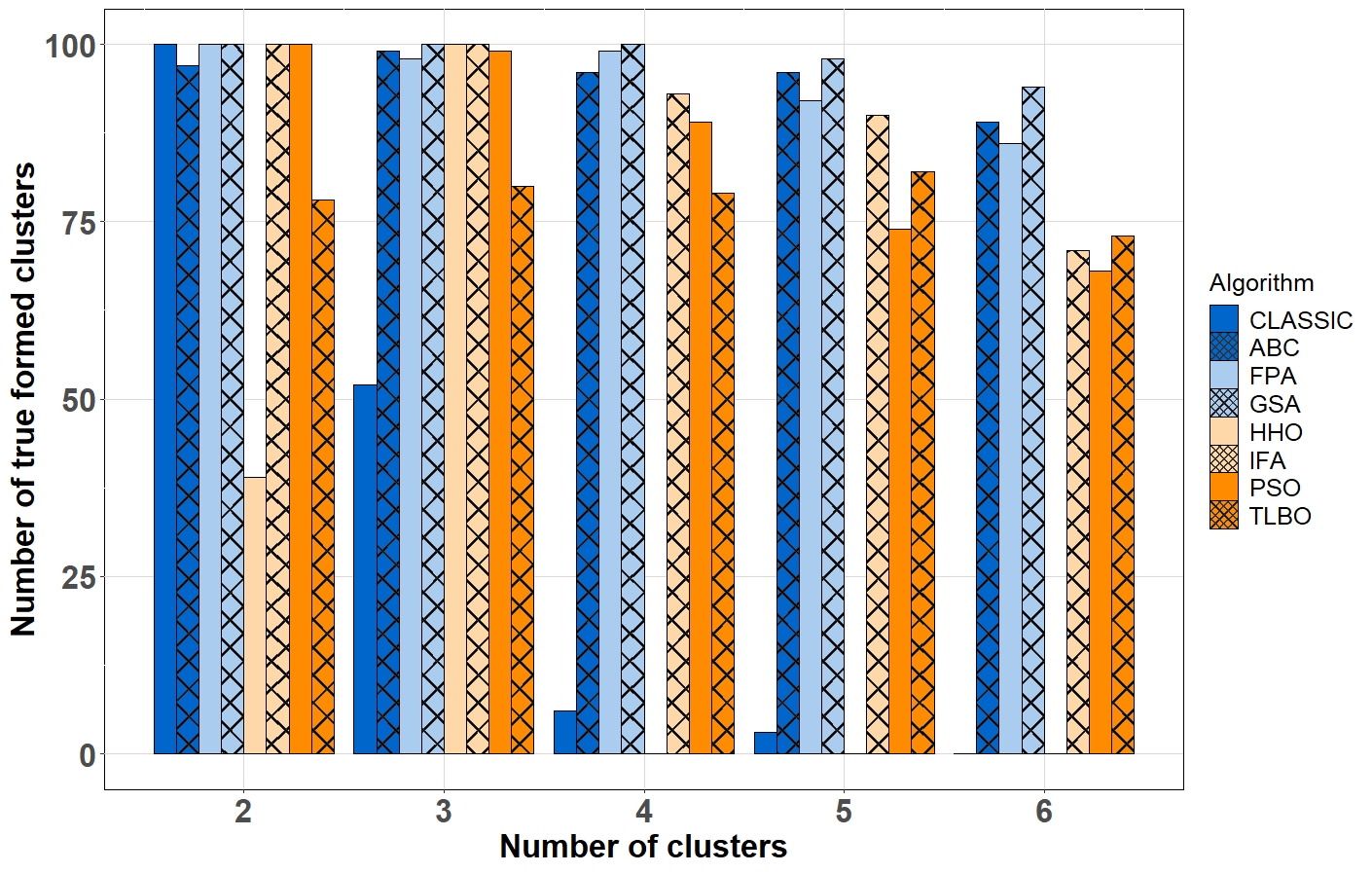}
        \caption{Case 3}
        \label{subfig:eval_3}
    \end{subfigure}
    \caption{Number of simulations with appropriate final number of clusters based on algorithm and number of clusters}
    \label{fig:eval_c}
\end{figure}

Figure~\ref{fig:eval_c} presents the number of successes in forming the final cluster output. Figures~\ref{subfig:eval_1}-\ref{subfig:eval_3} represent cases 1 to 3, respectively. From the figure, it is observed that as the specified number of clusters increases, the number of correctly formed clusters generally decreases, but the magnitude and pattern of this trend differ across algorithms and datasets. This variability indicates that each optimisation algorithm exhibits unique performance characteristics when optimising FGWC.

In Case 1, with a small and high dimensional datasets, all algorithms successfully formed two clusters, but performance diverged for three or more clusters (see Figure~\ref{subfig:eval_1}). Algorithms such as FPA, GSA, TLBO, and ABC consistently performed well with greater success, while the classic FGWC algorithm struggled beyond two clusters. In particular, the HHO algorithm exhibited inconsistent performance and failed significantly in clusters 4 and 6.

Meanwhile, Figure~\ref{subfig:eval_2} showed that most algorithms exhibited good performance in forming two to four clusters, but showed significant declines afterward. In contrast, ABC uniquely demonstrated an increasing trend of the number of sucesses. Based on the algorithms, it can be seen that GSA, FPA, PSO, and TLBO are the highest in each number of clusters. On the other hand, IFA only forms the high number of expected clusters only when the number of clusters is 2 and 3, and fails to produce 6 clusters. 

Based on Figure~\ref{subfig:eval_3}, it seems that in a larger dataset setting, all algorithms showed better overall consistency, with most algorithms maintaining a higher number of successes (above 75 out of 100 simulations) for cluster counts of two to six. However, the exception happens in classic FGWC and HHO. Classic FGWC performed well only at two clusters, while HHO performed well only up to three clusters and completely failed beyond that point.

In summary, the analysis indicates that optimisation algorithms significantly improve the performance of FGWC, particularly in correctly forming the desired number of clusters. Meanwhile, FPA, GSA, and TLBO emerge as the most reliable across all scenarios. IFA also tends to be reliable, but it showed failure in 6 clusters of cases 2. On the other hand, ABC shows potential in significantly different dimensions and observations (Cases 1 and 3), whereas PSO is preferable for cases with lower dimensions (Case 2 and 3). In contrast, HHO exhibits limited reliability, performing well only in case 2. Furthermore, the classic FGWC algorithm appears viable only in scenarios with balanced observations and dimensions, as demonstrated in Case 2.

\subsection{Performance evaluation} 
\label{ss:perf_eval}

Tables~\ref{tab:eval1}-\ref{tab:eval3c} summarise the evaluation results for each case according to fuzzy clustering criteria and computational cost. The bold values indicate the algorithm achieving optimal performance for each specific criterion and number of clusters. Evaluation metrics are calculated based on simulations that correctly formed the expected number of clusters; NA indicates that the algorithms that cannot form the expected clusters, as described in Section~\ref{sec:exp}.

Overall, results reveal that the classic FGWC algorithm generally produces the least optimal cluster separation, evident from the largest minimum index values and the smallest maximum index values. However, the classic FGWC consistently demonstrates the lowest computational time despite not having the lowest iteration count. In contrast, optimisation algorithms typically yield more optimal cluster separations compared to classic FGWC.

In Case 1, as presented in Table~\ref{tab:eval1}, TLBO consistently outperforms other algorithms in most clustering criteria (objective function, PC, CE, SC, and IFV indices), closely followed by ABC. However, TLBO and ABC also incur the highest computational costs, as shown in Table~\ref{tab:eval1c}. For the remaining indices, the optimal values vary by algorithm; for example, HHO achieves optimal values for the S and Kwon indices when it successfully forms clusters. Meanwhile, the XB index favors the classic FGWC for two clusters (c=2), ABC for six clusters (c=6), and IFA in other cluster counts. Therefore, in scenarios with significantly fewer observations compared to dimensions, TLBO and ABC exhibit strong clustering performance but at higher computational costs, as reflected by the results in Figure~\ref{subfig:eval_1}.

\begin{table}[ht]
\nointerlineskip\leavevmode
\centering
\caption{Performance evaluation summary for Case 1} 
\label{tab:eval1}
\resizebox{\textwidth}{!}{
\begin{tabular}{M{0.1\textwidth}rllllllll}
  \toprule
Criteria & c & ABC & CLASSIC & FPA & GSA & HHO & IFA & PSO & TLBO \\ 
  \midrule
\multirow{5}{*}{\shortstack[c]{Obj.\\function}} & 2 & $1.359\times10^{5}$ & $1.368\times10^{5}$ & $1.362\times10^{5}$ & $1.362\times10^{5}$ & $1.355\times10^{5}$ & $1.363\times10^{5}$ & $1.361\times10^{5}$ & $\boldsymbol{1.333\times10^{5}}$ \\ 
 & 3 & $1.103\times10^{5}$ & NA & $1.112\times10^{5}$ & $1.112\times10^{5}$ & $1.106\times10^{5}$ & $1.113\times10^{5}$ & $1.111\times10^{5}$ & $\boldsymbol{1.068\times10^{5}}$ \\ 
 & 4 & $9.263\times10^{4}$ & NA & $9.627\times10^{4}$ & $9.629\times10^{4}$ & NA & $9.637\times10^{4}$ & $9.628\times10^{4}$ & $\boldsymbol{8.847\times10^{4}}$ \\ 
 & 5 & $7.894\times10^{4}$ & NA & $8.615\times10^{4}$ & $8.623\times10^{4}$ & $8.591\times10^{4}$ & $8.627\times10^{4}$ & $8.537\times10^{4}$ & $\boldsymbol{7.638\times10^{4}}$ \\ 
 & 6 & $7.062\times10^{4}$ & NA & $7.859\times10^{4}$ & $7.863\times10^{4}$ & NA & $7.865\times10^{4}$ & $7.875\times10^{4}$ & $\boldsymbol{6.836\times10^{4}}$ \\ 
\multirow{5}{*}{\shortstack[c]{PC\\Index}} & 2 & $5.211\times10^{-1}$ & $5.000\times10^{-1}$ & $5.194\times10^{-1}$ & $5.193\times10^{-1}$ & $5.096\times10^{-1}$ & $5.108\times10^{-1}$ & $5.135\times10^{-1}$ & $\boldsymbol{5.878\times10^{-1}}$ \\ 
 & 3 & $3.563\times10^{-1}$ & NA & $3.530\times10^{-1}$ & $3.547\times10^{-1}$ & $3.558\times10^{-1}$ & $3.426\times10^{-1}$ & $3.525\times10^{-1}$ & $\boldsymbol{4.187\times10^{-1}}$ \\ 
 & 4 & $2.936\times10^{-1}$ & NA & $2.654\times10^{-1}$ & $2.656\times10^{-1}$ & NA & $2.562\times10^{-1}$ & $2.603\times10^{-1}$ & $\boldsymbol{3.444\times10^{-1}}$ \\ 
 & 5 & $2.443\times10^{-1}$ & NA & $2.118\times10^{-1}$ & $2.114\times10^{-1}$ & $2.163\times10^{-1}$ & $2.063\times10^{-1}$ & $2.080\times10^{-1}$ & $\boldsymbol{2.735\times10^{-1}}$ \\ 
 & 6 & $2.031\times10^{-1}$ & NA & $1.788\times10^{-1}$ & $1.785\times10^{-1}$ & NA & $1.783\times10^{-1}$ & $1.775\times10^{-1}$ & $\boldsymbol{2.231\times10^{-1}}$ \\ 
\multirow{5}{*}{\shortstack[c]{CE\\Index}} & 2 & $6.716\times10^{-1}$ & $6.931\times10^{-1}$ & $6.733\times10^{-1}$ & $6.734\times10^{-1}$ & $6.834\times10^{-1}$ & $6.823\times10^{-1}$ & $6.794\times10^{-1}$ & $\boldsymbol{5.986\times10^{-1}}$ \\ 
 & 3 & $1.061\times10^{0}$ & NA & $1.069\times10^{0}$ & $1.066\times10^{0}$ & $1.064\times10^{0}$ & $1.085\times10^{0}$ & $1.070\times10^{0}$ & $\boldsymbol{9.464\times10^{-1}}$ \\ 
 & 4 & $1.267\times10^{0}$ & NA & $1.355\times10^{0}$ & $1.355\times10^{0}$ & NA & $1.374\times10^{0}$ & $1.366\times10^{0}$ & $\boldsymbol{1.148\times10^{0}}$ \\ 
 & 5 & $1.449\times10^{0}$ & NA & $1.580\times10^{0}$ & $1.581\times10^{0}$ & $1.568\times10^{0}$ & $1.594\times10^{0}$ & $1.587\times10^{0}$ & $\boldsymbol{1.382\times10^{0}}$ \\ 
 & 6 & $1.638\times10^{0}$ & NA & $1.756\times10^{0}$ & $1.756\times10^{0}$ & NA & $1.757\times10^{0}$ & $1.761\times10^{0}$ & $\boldsymbol{1.583\times10^{0}}$ \\ 
\multirow{5}{*}{\shortstack[c]{SC\\Index}} & 2 & $1.203\times10^{2}$ & $6.284\times10^{6}$ & $1.432\times10^{2}$ & $1.285\times10^{2}$ & $3.252\times10^{1}$ & $1.848\times10^{2}$ & $1.380\times10^{2}$ & $\boldsymbol{3.056\times10^{1}}$ \\ 
 & 3 & $4.995\times10^{1}$ & NA & $6.368\times10^{1}$ & $5.813\times10^{1}$ & $4.376\times10^{1}$ & $1.199\times10^{2}$ & $5.001\times10^{1}$ & $\boldsymbol{7.249\times10^{0}}$ \\ 
 & 4 & $1.116\times10^{1}$ & NA & $5.030\times10^{1}$ & $5.095\times10^{1}$ & NA & $9.449\times10^{1}$ & $5.611\times10^{1}$ & $\boldsymbol{3.352\times10^{0}}$ \\ 
 & 5 & $5.109\times10^{0}$ & NA & $5.757\times10^{1}$ & $6.232\times10^{1}$ & $3.349\times10^{1}$ & $1.127\times10^{2}$ & $4.779\times10^{1}$ & $\boldsymbol{2.235\times10^{0}}$ \\ 
 & 6 & $3.987\times10^{0}$ & NA & $3.293\times10^{1}$ & $3.331\times10^{1}$ & NA & $3.871\times10^{1}$ & $2.741\times10^{1}$ & $\boldsymbol{1.778\times10^{0}}$ \\ 
\multirow{5}{*}{\shortstack[c]{S\\Index}} & 2 & $1.855\times10^{1}$ & $1.029\times10^{6}$ & $2.852\times10^{1}$ & $3.014\times10^{1}$ & $1.039\times10^{1}$ & $3.616\times10^{1}$ & $2.423\times10^{1}$ & $\boldsymbol{4.225\times10^{0}}$ \\ 
 & 3 & $8.076\times10^{2}$ & NA & $1.517\times10^{2}$ & $1.713\times10^{2}$ & $\boldsymbol{4.367\times10^{1}}$ & $4.163\times10^{2}$ & $2.354\times10^{3}$ & $2.050\times10^{2}$ \\ 
 & 4 & $3.866\times10^{3}$ & NA & $\boldsymbol{9.685\times10^{1}}$ & $9.986\times10^{1}$ & NA & $4.543\times10^{2}$ & $8.806\times10^{3}$ & $2.194\times10^{3}$ \\ 
 & 5 & $1.060\times10^{4}$ & NA & $4.512\times10^{2}$ & $5.227\times10^{2}$ & $\boldsymbol{5.731\times10^{1}}$ & $1.358\times10^{3}$ & $6.067\times10^{5}$ & $9.731\times10^{3}$ \\ 
 & 6 & $2.846\times10^{4}$ & NA & $7.850\times10^{1}$ & $6.220\times10^{1}$ & NA & $6.213\times10^{1}$ & $\boldsymbol{4.445\times10^{1}}$ & $8.705\times10^{3}$ \\ 
\multirow{5}{*}{\shortstack[c]{XB\\Index}} & 2 & $3.154\times10^{0}$ & $\boldsymbol{2.729\times10^{0}}$ & $3.087\times10^{0}$ & $3.078\times10^{0}$ & $3.183\times10^{0}$ & $3.031\times10^{0}$ & $3.108\times10^{0}$ & $3.405\times10^{0}$ \\ 
 & 3 & $2.628\times10^{0}$ & NA & $2.622\times10^{0}$ & $2.623\times10^{0}$ & $2.694\times10^{0}$ & $\boldsymbol{2.532\times10^{0}}$ & $2.664\times10^{0}$ & $2.910\times10^{0}$ \\ 
 & 4 & $2.348\times10^{0}$ & NA & $2.304\times10^{0}$ & $2.307\times10^{0}$ & NA & $\boldsymbol{2.169\times10^{0}}$ & $2.332\times10^{0}$ & $2.428\times10^{0}$ \\ 
 & 5 & $2.034\times10^{0}$ & NA & $2.053\times10^{0}$ & $2.044\times10^{0}$ & $2.073\times10^{0}$ & $\boldsymbol{1.946\times10^{0}}$ & $2.095\times10^{0}$ & $2.655\times10^{0}$ \\ 
 & 6 & $\boldsymbol{1.831\times10^{0}}$ & NA & $1.887\times10^{0}$ & $1.877\times10^{0}$ & NA & $1.855\times10^{0}$ & $1.861\times10^{0}$ & $3.082\times10^{0}$ \\ 
\multirow{5}{*}{\shortstack[c]{IFV\\Index}} & 2 & $7.602\times10^{-2}$ & $2.013\times10^{-6}$ & $6.117\times10^{-2}$ & $6.043\times10^{-2}$ & $9.844\times10^{-2}$ & $4.006\times10^{-2}$ & $5.322\times10^{-2}$ & $\boldsymbol{3.865\times10^{-1}}$ \\ 
 & 3 & $3.464\times10^{-1}$ & NA & $1.599\times10^{-1}$ & $1.645\times10^{-1}$ & $2.748\times10^{-1}$ & $9.541\times10^{-2}$ & $1.584\times10^{-1}$ & $\boldsymbol{2.088\times10^{0}}$ \\ 
 & 4 & $3.127\times10^{0}$ & NA & $2.143\times10^{-1}$ & $2.080\times10^{-1}$ & NA & $1.323\times10^{-1}$ & $1.688\times10^{-1}$ & $\boldsymbol{8.074\times10^{0}}$ \\ 
 & 5 & $8.525\times10^{0}$ & NA & $2.550\times10^{-1}$ & $1.923\times10^{-1}$ & $4.212\times10^{-1}$ & $1.434\times10^{-1}$ & $5.045\times10^{-1}$ & $\boldsymbol{1.111\times10^{1}}$ \\ 
 & 6 & $1.074\times10^{1}$ & NA & $3.639\times10^{-1}$ & $3.259\times10^{-1}$ & NA & $2.961\times10^{-1}$ & $2.498\times10^{-1}$ & $\boldsymbol{1.315\times10^{1}}$ \\ 
\multirow{5}{*}{\shortstack[c]{Kwon\\Index}} & 2 & $8.535\times10^{2}$ & $4.804\times10^{7}$ & $1.314\times10^{3}$ & $1.390\times10^{3}$ & $4.805\times10^{2}$ & $1.678\times10^{3}$ & $1.121\times10^{3}$ & $\boldsymbol{1.881\times10^{2}}$ \\ 
 & 3 & $4.522\times10^{4}$ & NA & $8.525\times10^{3}$ & $9.641\times10^{3}$ & $\boldsymbol{2.408\times10^{3}}$ & $2.362\times10^{4}$ & $1.335\times10^{5}$ & $1.110\times10^{4}$ \\ 
 & 4 & $2.389\times10^{5}$ & NA & $\boldsymbol{6.260\times10^{3}}$ & $6.457\times10^{3}$ & NA & $2.983\times10^{4}$ & $5.759\times10^{5}$ & $1.357\times10^{5}$ \\ 
 & 5 & $7.279\times10^{5}$ & NA & $3.305\times10^{4}$ & $3.827\times10^{4}$ & $\boldsymbol{4.051\times10^{3}}$ & $9.976\times10^{4}$ & $4.406\times10^{7}$ & $6.647\times10^{5}$ \\ 
 & 6 & $2.176\times10^{6}$ & NA & $6.147\times10^{3}$ & $4.845\times10^{3}$ & NA & $4.836\times10^{3}$ & $\boldsymbol{3.444\times10^{3}}$ & $6.487\times10^{5}$ \\ 
   \bottomrule
\end{tabular}
}
\end{table}

\begin{table}[ht]
\centering
\caption{Computational cost summary for Case 1} 
\label{tab:eval1c}
\resizebox{\textwidth}{!}{
\begin{tabular}{M{0.1\textwidth}rllllllll}
  \toprule
Criteria & c & ABC & CLASSIC & FPA & GSA & HHO & IFA & PSO & TLBO \\ 
  \midrule
\multirow{5}{*}{\shortstack[c]{Itera-\\tion}} & 2 & $1.399\times10^{1}$ & $1.924\times10^{1}$ & $1.065\times10^{1}$ & $\boldsymbol{1.058\times10^{1}}$ & $1.894\times10^{1}$ & $1.280\times10^{1}$ & $1.737\times10^{1}$ & $2.370\times10^{1}$ \\ 
 & 3 & $2.000\times10^{1}$ & NA & $1.038\times10^{1}$ & $\boldsymbol{1.014\times10^{1}}$ & $1.707\times10^{1}$ & $1.261\times10^{1}$ & $1.251\times10^{1}$ & $2.714\times10^{1}$ \\ 
 & 4 & $3.703\times10^{1}$ & NA & $1.039\times10^{1}$ & $\boldsymbol{1.031\times10^{1}}$ & NA & $1.267\times10^{1}$ & $1.504\times10^{1}$ & $3.048\times10^{1}$ \\ 
 & 5 & $4.331\times10^{1}$ & NA & $1.047\times10^{1}$ & $1.037\times10^{1}$ & $\boldsymbol{1.000\times10^{1}}$ & $1.238\times10^{1}$ & $2.900\times10^{1}$ & $3.205\times10^{1}$ \\ 
 & 6 & $4.422\times10^{1}$ & NA & $1.007\times10^{1}$ & $\boldsymbol{1.000\times10^{1}}$ & NA & $1.100\times10^{1}$ & $1.000\times10^{1}$ & $3.211\times10^{1}$ \\ 
\multirow{5}{*}{\shortstack[c]{Comp.\\time}} & 2 & $1.081\times10^{0}$ & $\boldsymbol{3.260\times10^{-2}}$ & $1.510\times10^{-1}$ & $2.922\times10^{-1}$ & $3.050\times10^{-1}$ & $7.712\times10^{-1}$ & $6.623\times10^{-1}$ & $1.277\times10^{0}$ \\ 
 & 3 & $1.678\times10^{0}$ & NA & $\boldsymbol{2.458\times10^{-1}}$ & $6.091\times10^{-1}$ & $5.424\times10^{-1}$ & $1.110\times10^{0}$ & $2.833\times10^{-1}$ & $1.023\times10^{0}$ \\ 
 & 4 & $4.302\times10^{0}$ & NA & $\boldsymbol{2.268\times10^{-1}}$ & $3.592\times10^{-1}$ & NA & $5.353\times10^{-1}$ & $3.079\times10^{-1}$ & $1.166\times10^{0}$ \\ 
 & 5 & $5.227\times10^{0}$ & NA & $2.136\times10^{-1}$ & $3.893\times10^{-1}$ & $\boldsymbol{1.755\times10^{-1}}$ & $5.994\times10^{-1}$ & $5.500\times10^{-1}$ & $1.258\times10^{0}$ \\ 
 & 6 & $5.452\times10^{0}$ & NA & $2.504\times10^{-1}$ & $4.340\times10^{-1}$ & NA & $5.980\times10^{-1}$ & $\boldsymbol{2.329\times10^{-1}}$ & $1.471\times10^{0}$ \\ 
   \bottomrule
\end{tabular}
}
\end{table}

Based on Table~\ref{tab:eval2}, in Case 2, FGWC using ABC and HHO consistently achieve optimal cluster separations across all clustering criteria, except XB. TLBO closely follows ABC and HHO, indicating strong performance across these criteria. The performance gap between these three algorithms (ABC, HHO, and TLBO) and other algorithms (FPA, GSA, IFA, and PSO) is relatively large. However, the XB index shows variable results, with different algorithms performing optimally for different cluster counts. From a computational point of view (Table~\ref{tab:eval2c}), FPA and PSO require the fewest iterations, indicating lower computational cost. However, despite the strong clustering results of ABC and HHO, their lower frequency to form the expected number of clusters (Figure~\ref{subfig:eval_2}) reduces the generalizability of their performance advantage over TLBO. Therefore, TLBO demonstrates promising results, considering both the cluster evaluation measurements and the frequency of the expected simulations.

\begin{table}[ht]
\centering
\caption{Performance evaluation summary for Case 2} 
\label{tab:eval2}
\resizebox{\textwidth}{!}{
\begin{tabular}{M{0.1\textwidth}rllllllll}
  \toprule
Criteria & c & ABC & CLASSIC & FPA & GSA & HHO & IFA & PSO & TLBO \\ 
  \midrule
\multirow{5}{*}{\shortstack[c]{Obj.\\function}} & 2 & $\boldsymbol{4.727\times10^{4}}$ & $5.809\times10^{4}$ & $5.691\times10^{4}$ & $5.778\times10^{4}$ & $4.942\times10^{4}$ & $5.789\times10^{4}$ & $5.707\times10^{4}$ & $4.891\times10^{4}$ \\ 
 & 3 & $\boldsymbol{3.702\times10^{4}}$ & $4.635\times10^{4}$ & $4.493\times10^{4}$ & $3.929\times10^{4}$ & $3.810\times10^{4}$ & $4.120\times10^{4}$ & $3.946\times10^{4}$ & $3.810\times10^{4}$ \\ 
 & 4 & $\boldsymbol{2.497\times10^{4}}$ & $3.793\times10^{4}$ & $3.465\times10^{4}$ & $3.357\times10^{4}$ & $3.011\times10^{4}$ & $3.619\times10^{4}$ & $3.645\times10^{4}$ & $3.179\times10^{4}$ \\ 
 & 5 & $2.029\times10^{4}$ & $3.247\times10^{4}$ & $2.910\times10^{4}$ & $3.030\times10^{4}$ & $\boldsymbol{2.007\times10^{4}}$ & $3.141\times10^{4}$ & $3.247\times10^{4}$ & $2.646\times10^{4}$ \\ 
 & 6 & $\boldsymbol{1.771\times10^{4}}$ & $3.036\times10^{4}$ & $2.806\times10^{4}$ & $2.828\times10^{4}$ & $2.091\times10^{4}$ & NA & $3.249\times10^{4}$ & $2.370\times10^{4}$ \\ 
\multirow{5}{*}{\shortstack[c]{PC\\Index}} & 2 & $\boldsymbol{7.869\times10^{-1}}$ & $5.000\times10^{-1}$ & $5.512\times10^{-1}$ & $5.198\times10^{-1}$ & $7.347\times10^{-1}$ & $5.136\times10^{-1}$ & $5.523\times10^{-1}$ & $7.482\times10^{-1}$ \\ 
 & 3 & $4.556\times10^{-1}$ & $4.334\times10^{-1}$ & $3.703\times10^{-1}$ & $4.059\times10^{-1}$ & $\boldsymbol{5.228\times10^{-1}}$ & $3.907\times10^{-1}$ & $4.046\times10^{-1}$ & $4.267\times10^{-1}$ \\ 
 & 4 & $\boldsymbol{4.028\times10^{-1}}$ & $3.637\times10^{-1}$ & $3.117\times10^{-1}$ & $2.871\times10^{-1}$ & $3.696\times10^{-1}$ & $2.768\times10^{-1}$ & $2.784\times10^{-1}$ & $3.178\times10^{-1}$ \\ 
 & 5 & $3.080\times10^{-1}$ & $2.733\times10^{-1}$ & $2.488\times10^{-1}$ & $2.233\times10^{-1}$ & $\boldsymbol{3.390\times10^{-1}}$ & $2.210\times10^{-1}$ & $2.187\times10^{-1}$ & $2.594\times10^{-1}$ \\ 
 & 6 & $2.504\times10^{-1}$ & $1.973\times10^{-1}$ & $1.974\times10^{-1}$ & $1.833\times10^{-1}$ & $\boldsymbol{2.640\times10^{-1}}$ & NA & $1.778\times10^{-1}$ & $2.138\times10^{-1}$ \\ 
\multirow{5}{*}{\shortstack[c]{CE\\Index}} & 2 & $\boldsymbol{3.578\times10^{-1}}$ & $6.931\times10^{-1}$ & $6.384\times10^{-1}$ & $6.731\times10^{-1}$ & $4.271\times10^{-1}$ & $6.794\times10^{-1}$ & $6.394\times10^{-1}$ & $4.076\times10^{-1}$ \\ 
 & 3 & $8.903\times10^{-1}$ & $8.909\times10^{-1}$ & $1.033\times10^{0}$ & $9.650\times10^{-1}$ & $\boldsymbol{8.139\times10^{-1}}$ & $9.937\times10^{-1}$ & $9.676\times10^{-1}$ & $9.338\times10^{-1}$ \\ 
 & 4 & $\boldsymbol{1.011\times10^{0}}$ & $1.093\times10^{0}$ & $1.233\times10^{0}$ & $1.297\times10^{0}$ & $1.125\times10^{0}$ & $1.323\times10^{0}$ & $1.321\times10^{0}$ & $1.233\times10^{0}$ \\ 
 & 5 & $1.276\times10^{0}$ & $1.361\times10^{0}$ & $1.452\times10^{0}$ & $1.544\times10^{0}$ & $\boldsymbol{1.221\times10^{0}}$ & $1.551\times10^{0}$ & $1.558\times10^{0}$ & $1.437\times10^{0}$ \\ 
 & 6 & $1.488\times10^{0}$ & $1.678\times10^{0}$ & $1.677\times10^{0}$ & $1.740\times10^{0}$ & $\boldsymbol{1.486\times10^{0}}$ & NA & $1.758\times10^{0}$ & $1.631\times10^{0}$ \\ 
\multirow{5}{*}{\shortstack[c]{SC\\Index}} & 2 & $2.034\times10^{1}$ & $1.445\times10^{8}$ & $1.738\times10^{2}$ & $8.384\times10^{2}$ & $\boldsymbol{1.324\times10^{1}}$ & $8.610\times10^{2}$ & $3.017\times10^{2}$ & $2.572\times10^{1}$ \\ 
 & 3 & $3.663\times10^{0}$ & $3.045\times10^{0}$ & $7.588\times10^{1}$ & $9.974\times10^{0}$ & $4.718\times10^{0}$ & $7.412\times10^{1}$ & $7.310\times10^{0}$ & $\boldsymbol{2.965\times10^{0}}$ \\ 
 & 4 & $\boldsymbol{1.226\times10^{0}}$ & $1.509\times10^{0}$ & $3.064\times10^{1}$ & $2.531\times10^{0}$ & $1.792\times10^{0}$ & $5.500\times10^{1}$ & $2.586\times10^{1}$ & $2.143\times10^{0}$ \\ 
 & 5 & $8.411\times10^{-1}$ & $1.063\times10^{0}$ & $2.203\times10^{1}$ & $4.198\times10^{0}$ & $\boldsymbol{7.987\times10^{-1}}$ & $1.702\times10^{1}$ & $2.362\times10^{1}$ & $1.573\times10^{0}$ \\ 
 & 6 & $\boldsymbol{6.789\times10^{-1}}$ & $1.517\times10^{0}$ & $2.696\times10^{1}$ & $8.310\times10^{0}$ & $1.483\times10^{0}$ & NA & $3.788\times10^{1}$ & $1.349\times10^{0}$ \\ 
\multirow{5}{*}{\shortstack[c]{S\\Index}} & 2 & $\boldsymbol{5.056\times10^{-1}}$ & $5.305\times10^{6}$ & $2.491\times10^{1}$ & $3.346\times10^{1}$ & $7.267\times10^{-1}$ & $4.513\times10^{1}$ & $8.935\times10^{0}$ & $8.921\times10^{-1}$ \\ 
 & 3 & $8.392\times10^{1}$ & $1.451\times10^{24}$ & $5.446\times10^{14}$ & $5.044\times10^{20}$ & $\boldsymbol{5.395\times10^{0}}$ & $3.110\times10^{21}$ & $2.234\times10^{5}$ & $1.183\times10^{11}$ \\ 
 & 4 & $1.388\times10^{3}$ & $8.908\times10^{15}$ & $2.165\times10^{10}$ & $3.791\times10^{16}$ & $\boldsymbol{7.483\times10^{0}}$ & $6.003\times10^{17}$ & $1.903\times10^{5}$ & $1.624\times10^{6}$ \\ 
 & 5 & $1.894\times10^{3}$ & $2.218\times10^{11}$ & $6.284\times10^{6}$ & $1.287\times10^{13}$ & $\boldsymbol{7.739\times10^{0}}$ & $3.419\times10^{17}$ & $9.072\times10^{4}$ & $1.410\times10^{5}$ \\ 
 & 6 & $1.660\times10^{3}$ & $5.205\times10^{16}$ & $5.885\times10^{8}$ & $2.455\times10^{13}$ & $\boldsymbol{1.191\times10^{1}}$ & NA & $3.578\times10^{5}$ & $2.373\times10^{4}$ \\ 
\multirow{5}{*}{\shortstack[c]{XB\\Index}} & 2 & $2.478\times10^{1}$ & $\boldsymbol{8.061\times10^{0}}$ & $1.321\times10^{1}$ & $1.134\times10^{1}$ & $2.695\times10^{1}$ & $1.059\times10^{1}$ & $1.453\times10^{1}$ & $2.529\times10^{1}$ \\ 
 & 3 & $2.197\times10^{1}$ & $1.375\times10^{1}$ & $\boldsymbol{1.233\times10^{1}}$ & $1.663\times10^{1}$ & $2.238\times10^{1}$ & $1.473\times10^{1}$ & $1.668\times10^{1}$ & $2.062\times10^{1}$ \\ 
 & 4 & $3.030\times10^{1}$ & $2.195\times10^{1}$ & $1.555\times10^{1}$ & $1.329\times10^{1}$ & $2.518\times10^{1}$ & $\boldsymbol{1.152\times10^{1}}$ & $1.249\times10^{1}$ & $2.033\times10^{1}$ \\ 
 & 5 & $3.102\times10^{1}$ & $2.354\times10^{1}$ & $1.750\times10^{1}$ & $1.101\times10^{1}$ & $3.154\times10^{1}$ & $\boldsymbol{1.026\times10^{1}}$ & $1.056\times10^{1}$ & $2.308\times10^{1}$ \\ 
 & 6 & $3.033\times10^{1}$ & $1.504\times10^{1}$ & $1.453\times10^{1}$ & $\boldsymbol{9.389\times10^{0}}$ & $2.027\times10^{1}$ & NA & $1.038\times10^{1}$ & $2.164\times10^{1}$ \\ 
\multirow{5}{*}{\shortstack[c]{IFV\\Index}} & 2 & $\boldsymbol{1.473\times10^{0}}$ & $1.908\times10^{-7}$ & $1.506\times10^{-1}$ & $3.477\times10^{-2}$ & $1.155\times10^{0}$ & $2.534\times10^{-2}$ & $1.140\times10^{-1}$ & $1.240\times10^{0}$ \\ 
 & 3 & $3.947\times10^{0}$ & $\boldsymbol{4.564\times10^{0}}$ & $1.040\times10^{0}$ & $2.709\times10^{0}$ & $3.316\times10^{0}$ & $2.076\times10^{0}$ & $2.670\times10^{0}$ & $3.335\times10^{0}$ \\ 
 & 4 & $\boldsymbol{1.370\times10^{1}}$ & $1.154\times10^{1}$ & $5.492\times10^{0}$ & $3.824\times10^{0}$ & $8.272\times10^{0}$ & $2.492\times10^{0}$ & $2.384\times10^{0}$ & $5.706\times10^{0}$ \\ 
 & 5 & $2.027\times10^{1}$ & $1.690\times10^{1}$ & $9.572\times10^{0}$ & $4.265\times10^{0}$ & $\boldsymbol{2.227\times10^{1}}$ & $3.539\times10^{0}$ & $2.857\times10^{0}$ & $1.020\times10^{1}$ \\ 
 & 6 & $\boldsymbol{2.498\times10^{1}}$ & $1.091\times10^{1}$ & $8.723\times10^{0}$ & $4.212\times10^{0}$ & $2.047\times10^{1}$ & NA & $8.800\times10^{-1}$ & $1.257\times10^{1}$ \\ 
\multirow{5}{*}{\shortstack[c]{Kwon\\Index}} & 2 & $\boldsymbol{2.528\times10^{1}}$ & $2.851\times10^{8}$ & $1.308\times10^{3}$ & $1.783\times10^{3}$ & $3.750\times10^{1}$ & $2.409\times10^{3}$ & $4.723\times10^{2}$ & $4.597\times10^{1}$ \\ 
 & 3 & $5.687\times10^{3}$ & $1.002\times10^{26}$ & $3.758\times10^{16}$ & $3.480\times10^{22}$ & $\boldsymbol{3.430\times10^{2}}$ & $2.146\times10^{23}$ & $1.543\times10^{7}$ & $8.164\times10^{12}$ \\ 
 & 4 & $1.343\times10^{5}$ & $9.360\times10^{17}$ & $2.247\times10^{12}$ & $3.140\times10^{18}$ & $\boldsymbol{6.330\times10^{2}}$ & $4.972\times10^{19}$ & $1.577\times10^{7}$ & $1.345\times10^{8}$ \\ 
 & 5 & $2.492\times10^{5}$ & $2.148\times10^{13}$ & $7.546\times10^{8}$ & $1.214\times10^{15}$ & $\boldsymbol{9.535\times10^{2}}$ & $3.225\times10^{19}$ & $8.562\times10^{6}$ & $1.333\times10^{7}$ \\ 
 & 6 & $2.565\times10^{5}$ & $5.435\times10^{18}$ & $6.145\times10^{10}$ & $2.563\times10^{15}$ & $\boldsymbol{1.558\times10^{3}}$ & NA & $3.736\times10^{7}$ & $2.521\times10^{6}$ \\ 
   \bottomrule
\end{tabular}
}
\end{table}

\begin{table}[ht]
\centering
\caption{Computational cost summary for Case 2} 
\label{tab:eval2c}
\resizebox{\textwidth}{!}{
\begin{tabular}{M{0.1\textwidth}rllllllll}
  \toprule
Criteria & c & ABC & CLASSIC & FPA & GSA & HHO & IFA & PSO & TLBO \\ 
  \midrule
\multirow{5}{*}{\shortstack[c]{Itera-\\tion}} & 2 & $2.300\times10^{1}$ & $1.875\times10^{2}$ & $\boldsymbol{1.201\times10^{1}}$ & $1.742\times10^{1}$ & $2.034\times10^{1}$ & $1.268\times10^{1}$ & $3.573\times10^{1}$ & $2.373\times10^{1}$ \\ 
 & 3 & $2.181\times10^{1}$ & $9.023\times10^{1}$ & $\boldsymbol{1.767\times10^{1}}$ & $5.814\times10^{1}$ & $2.271\times10^{1}$ & $5.098\times10^{1}$ & $3.233\times10^{1}$ & $2.347\times10^{1}$ \\ 
 & 4 & $3.626\times10^{1}$ & $2.395\times10^{1}$ & $2.289\times10^{1}$ & $4.423\times10^{1}$ & $2.709\times10^{1}$ & $3.287\times10^{1}$ & $\boldsymbol{2.194\times10^{1}}$ & $2.312\times10^{1}$ \\ 
 & 5 & $3.466\times10^{1}$ & $2.227\times10^{1}$ & $\boldsymbol{2.226\times10^{1}}$ & $4.062\times10^{1}$ & $2.899\times10^{1}$ & $4.060\times10^{1}$ & $2.300\times10^{1}$ & $2.376\times10^{1}$ \\ 
 & 6 & $3.736\times10^{1}$ & $3.067\times10^{1}$ & $2.044\times10^{1}$ & $4.061\times10^{1}$ & $2.760\times10^{1}$ & NA & $\boldsymbol{1.308\times10^{1}}$ & $2.430\times10^{1}$ \\ 
\multirow{5}{*}{\shortstack[c]{Comp.\\time}} & 2 & $6.250\times10^{-1}$ & $1.227\times10^{-1}$ & $\boldsymbol{1.070\times10^{-1}}$ & $3.192\times10^{-1}$ & $1.734\times10^{-1}$ & $2.912\times10^{-1}$ & $3.173\times10^{-1}$ & $5.271\times10^{-1}$ \\ 
 & 3 & $7.367\times10^{-1}$ & $\boldsymbol{7.400\times10^{-2}}$ & $1.756\times10^{-1}$ & $1.108\times10^{0}$ & $2.004\times10^{-1}$ & $1.174\times10^{0}$ & $2.775\times10^{-1}$ & $5.086\times10^{-1}$ \\ 
 & 4 & $1.356\times10^{0}$ & $\boldsymbol{2.795\times10^{-2}}$ & $2.558\times10^{-1}$ & $9.150\times10^{-1}$ & $2.660\times10^{-1}$ & $8.548\times10^{-1}$ & $2.129\times10^{-1}$ & $5.698\times10^{-1}$ \\ 
 & 5 & $1.506\times10^{0}$ & $\boldsymbol{3.267\times10^{-2}}$ & $2.743\times10^{-1}$ & $9.009\times10^{-1}$ & $2.918\times10^{-1}$ & $1.187\times10^{0}$ & $2.469\times10^{-1}$ & $5.134\times10^{-1}$ \\ 
 & 6 & $1.792\times10^{0}$ & $\boldsymbol{4.333\times10^{-2}}$ & $2.778\times10^{-1}$ & $8.633\times10^{-1}$ & $3.068\times10^{-1}$ & NA & $1.296\times10^{-1}$ & $5.614\times10^{-1}$ \\ 
   \bottomrule
\end{tabular}
}
\end{table}

Similarly, Table~\ref{tab:eval3} shows a pattern similar to Table~\ref{tab:eval1}, with TLBO and ABC again performing strongly in most evaluation metrics. However, unlike Case 1, the superiority between TLBO and ABC varies more between different cluster counts, with ABC occasionally surpassing TLBO. Additionally, HHO performs well only at lower number of clusters, but fails at higher numbers (four and above). Algorithms such as GSA, PSO, FPA, and IFA offer the next best cluster separations consistently. Interestingly, classic FGWC maintains the lowest XB index across clusters, except when it is unable to produce the expected clusters (e.g., six clusters). Computationally, the classic FGWC again records the lowest time, while FPA and GSA have fewer iterations (Table~\ref{tab:eval3c}). However, TLBO and ABC are among those that require the most iterations. Integrating these results with Figure~\ref{subfig:eval_3}, all optimisation algorithms except HHO perform well when observations significantly exceed dimensions, particularly ABC and TLBO.

In conclusion, while optimisation algorithms effectively address weaknesses in classic FGWC, each algorithm having specific advantages depending on the characteristics of the dataset. However, the main limitation of optimisation algorithms is their computational cost, approximately ten times higher than that of the classic FGWC, due to population-based searches. Among the algorithms tested, TLBO consistently produces superior clustering results across all cases, benefiting from a higher number of iterations that prevent convergence to local optima, a common issue in other optimisation algorithms. Consequently, further development to improve the computational efficiency of these algorithms is essential.

\begin{table}[ht]
\centering
\caption{Performance evaluation summary for Case 3} 
\label{tab:eval3}
\resizebox{\textwidth}{!}{
\begin{tabular}{M{0.1\textwidth}rllllllll}
  \toprule
Criteria & c & ABC & CLASSIC & FPA & GSA & HHO & IFA & PSO & TLBO \\ 
  \midrule
\multirow{5}{*}{\shortstack[c]{Obj.\\function}} & 2 & $4.651\times10^{5}$ & $4.669\times10^{5}$ & $4.668\times10^{5}$ & $4.669\times10^{5}$ & $4.647\times10^{5}$ & $4.669\times10^{5}$ & $4.664\times10^{5}$ & $\boldsymbol{4.631\times10^{5}}$ \\ 
 & 3 & $\boldsymbol{3.704\times10^{5}}$ & $3.813\times10^{5}$ & $3.812\times10^{5}$ & $3.812\times10^{5}$ & $3.769\times10^{5}$ & $3.812\times10^{5}$ & $3.808\times10^{5}$ & $3.725\times10^{5}$ \\ 
 & 4 & $\boldsymbol{3.160\times10^{5}}$ & $3.302\times10^{5}$ & $3.301\times10^{5}$ & $3.301\times10^{5}$ & NA & $3.301\times10^{5}$ & $3.299\times10^{5}$ & $3.170\times10^{5}$ \\ 
 & 5 & $\boldsymbol{2.673\times10^{5}}$ & $2.953\times10^{5}$ & $2.949\times10^{5}$ & $2.953\times10^{5}$ & NA & $2.953\times10^{5}$ & $2.950\times10^{5}$ & $2.709\times10^{5}$ \\ 
 & 6 & $\boldsymbol{2.399\times10^{5}}$ & NA & $2.691\times10^{5}$ & $2.695\times10^{5}$ & NA & $2.695\times10^{5}$ & $2.693\times10^{5}$ & $2.470\times10^{5}$ \\ 
\multirow{5}{*}{\shortstack[c]{PC\\Index}} & 2 & $5.135\times10^{-1}$ & $5.000\times10^{-1}$ & $5.136\times10^{-1}$ & $5.151\times10^{-1}$ & $5.256\times10^{-1}$ & $5.030\times10^{-1}$ & $5.020\times10^{-1}$ & $\boldsymbol{5.384\times10^{-1}}$ \\ 
 & 3 & $3.710\times10^{-1}$ & $3.333\times10^{-1}$ & $3.452\times10^{-1}$ & $3.463\times10^{-1}$ & $\boldsymbol{4.011\times10^{-1}}$ & $3.356\times10^{-1}$ & $3.347\times10^{-1}$ & $3.827\times10^{-1}$ \\ 
 & 4 & $2.768\times10^{-1}$ & $2.500\times10^{-1}$ & $2.588\times10^{-1}$ & $2.602\times10^{-1}$ & NA & $2.518\times10^{-1}$ & $2.511\times10^{-1}$ & $\boldsymbol{2.937\times10^{-1}}$ \\ 
 & 5 & $2.374\times10^{-1}$ & $2.000\times10^{-1}$ & $2.080\times10^{-1}$ & $2.085\times10^{-1}$ & NA & $2.020\times10^{-1}$ & $2.007\times10^{-1}$ & $\boldsymbol{2.447\times10^{-1}}$ \\ 
 & 6 & $\boldsymbol{1.972\times10^{-1}}$ & NA & $1.729\times10^{-1}$ & $1.744\times10^{-1}$ & NA & $1.685\times10^{-1}$ & $1.673\times10^{-1}$ & $1.965\times10^{-1}$ \\ 
\multirow{5}{*}{\shortstack[c]{CE\\Index}} & 2 & $6.793\times10^{-1}$ & $6.931\times10^{-1}$ & $6.794\times10^{-1}$ & $6.778\times10^{-1}$ & $6.670\times10^{-1}$ & $6.901\times10^{-1}$ & $6.912\times10^{-1}$ & $\boldsymbol{6.534\times10^{-1}}$ \\ 
 & 3 & $1.035\times10^{0}$ & $1.099\times10^{0}$ & $1.081\times10^{0}$ & $1.079\times10^{0}$ & $\boldsymbol{1.003\times10^{0}}$ & $1.095\times10^{0}$ & $1.097\times10^{0}$ & $1.021\times10^{0}$ \\ 
 & 4 & $1.325\times10^{0}$ & $1.386\times10^{0}$ & $1.369\times10^{0}$ & $1.366\times10^{0}$ & NA & $1.383\times10^{0}$ & $1.384\times10^{0}$ & $\boldsymbol{1.281\times10^{0}}$ \\ 
 & 5 & $1.495\times10^{0}$ & $1.609\times10^{0}$ & $1.589\times10^{0}$ & $1.588\times10^{0}$ & NA & $1.605\times10^{0}$ & $1.608\times10^{0}$ & $\boldsymbol{1.459\times10^{0}}$ \\ 
 & 6 & $1.677\times10^{0}$ & NA & $1.772\times10^{0}$ & $1.769\times10^{0}$ & NA & $1.786\times10^{0}$ & $1.790\times10^{0}$ & $\boldsymbol{1.673\times10^{0}}$ \\ 
\multirow{5}{*}{\shortstack[c]{SC\\Index}} & 2 & $4.207\times10^{2}$ & $4.856\times10^{6}$ & $1.700\times10^{3}$ & $1.883\times10^{3}$ & $\boldsymbol{2.007\times10^{2}}$ & $2.810\times10^{3}$ & $3.651\times10^{2}$ & $1.392\times10^{3}$ \\ 
 & 3 & $\boldsymbol{4.977\times10^{1}}$ & $3.508\times10^{6}$ & $8.684\times10^{2}$ & $1.092\times10^{3}$ & $1.093\times10^{2}$ & $1.580\times10^{3}$ & $2.240\times10^{2}$ & $7.137\times10^{1}$ \\ 
 & 4 & $\boldsymbol{3.184\times10^{1}}$ & $2.097\times10^{7}$ & $7.730\times10^{2}$ & $8.341\times10^{2}$ & NA & $1.340\times10^{3}$ & $2.521\times10^{2}$ & $4.216\times10^{1}$ \\ 
 & 5 & $\boldsymbol{6.829\times10^{0}}$ & $2.859\times10^{6}$ & $5.875\times10^{2}$ & $6.349\times10^{2}$ & NA & $1.145\times10^{3}$ & $2.113\times10^{2}$ & $1.424\times10^{1}$ \\ 
 & 6 & $\boldsymbol{6.639\times10^{0}}$ & NA & $5.744\times10^{2}$ & $6.083\times10^{2}$ & NA & $9.588\times10^{2}$ & $2.000\times10^{2}$ & $2.904\times10^{1}$ \\ 
\multirow{5}{*}{\shortstack[c]{S\\Index}} & 2 & $1.019\times10^{2}$ & $1.643\times10^{6}$ & $5.259\times10^{2}$ & $5.384\times10^{2}$ & $\boldsymbol{1.768\times10^{1}}$ & $8.370\times10^{2}$ & $1.133\times10^{2}$ & $2.119\times10^{1}$ \\ 
 & 3 & $3.364\times10^{2}$ & $1.013\times10^{11}$ & $1.085\times10^{3}$ & $1.380\times10^{3}$ & $\boldsymbol{8.699\times10^{0}}$ & $1.723\times10^{3}$ & $1.764\times10^{3}$ & $4.395\times10^{1}$ \\ 
 & 4 & $2.935\times10^{3}$ & $1.914\times10^{10}$ & $1.929\times10^{3}$ & $1.754\times10^{3}$ & NA & $3.739\times10^{3}$ & $1.813\times10^{3}$ & $\boldsymbol{5.666\times10^{2}}$ \\ 
 & 5 & $2.230\times10^{3}$ & $6.748\times10^{11}$ & $9.802\times10^{6}$ & $2.474\times10^{3}$ & NA & $4.571\times10^{3}$ & $9.039\times10^{3}$ & $\boldsymbol{1.059\times10^{3}}$ \\ 
 & 6 & $2.001\times10^{3}$ & NA & $2.268\times10^{3}$ & $\boldsymbol{1.876\times10^{3}}$ & NA & $3.860\times10^{3}$ & $5.616\times10^{3}$ & $3.694\times10^{3}$ \\ 
\multirow{5}{*}{\shortstack[c]{XB\\Index}} & 2 & $1.214\times10^{1}$ & $\boldsymbol{1.129\times10^{1}}$ & $1.177\times10^{1}$ & $1.177\times10^{1}$ & $1.189\times10^{1}$ & $1.154\times10^{1}$ & $1.169\times10^{1}$ & $1.234\times10^{1}$ \\ 
 & 3 & $1.010\times10^{1}$ & $\boldsymbol{9.225\times10^{0}}$ & $9.761\times10^{0}$ & $9.728\times10^{0}$ & $1.160\times10^{1}$ & $9.490\times10^{0}$ & $9.530\times10^{0}$ & $1.069\times10^{1}$ \\ 
 & 4 & $8.698\times10^{0}$ & $\boldsymbol{7.986\times10^{0}}$ & $8.462\times10^{0}$ & $8.491\times10^{0}$ & NA & $8.235\times10^{0}$ & $8.372\times10^{0}$ & $9.271\times10^{0}$ \\ 
 & 5 & $7.834\times10^{0}$ & $\boldsymbol{7.144\times10^{0}}$ & $7.576\times10^{0}$ & $7.581\times10^{0}$ & NA & $7.387\times10^{0}$ & $7.474\times10^{0}$ & $7.977\times10^{0}$ \\ 
 & 6 & $7.136\times10^{0}$ & NA & $6.934\times10^{0}$ & $6.950\times10^{0}$ & NA & $\boldsymbol{6.770\times10^{0}}$ & $6.903\times10^{0}$ & $7.198\times10^{0}$ \\ 
\multirow{5}{*}{\shortstack[c]{IFV\\Index}} & 2 & $4.624\times10^{-2}$ & $2.610\times10^{-6}$ & $3.336\times10^{-3}$ & $3.277\times10^{-3}$ & $6.315\times10^{-2}$ & $1.845\times10^{-3}$ & $1.024\times10^{-2}$ & $\boldsymbol{1.059\times10^{-1}}$ \\ 
 & 3 & $7.838\times10^{-1}$ & $7.684\times10^{-6}$ & $8.301\times10^{-3}$ & $7.328\times10^{-3}$ & $2.578\times10^{-1}$ & $4.454\times10^{-3}$ & $2.459\times10^{-2}$ & $\boldsymbol{8.618\times10^{-1}}$ \\ 
 & 4 & $1.670\times10^{0}$ & $1.076\times10^{-6}$ & $1.022\times10^{-2}$ & $9.387\times10^{-3}$ & NA & $5.752\times10^{-3}$ & $2.959\times10^{-2}$ & $\boldsymbol{3.337\times10^{0}}$ \\ 
 & 5 & $6.129\times10^{0}$ & $1.411\times10^{-6}$ & $1.565\times10^{-1}$ & $1.256\times10^{-2}$ & NA & $7.636\times10^{-3}$ & $3.407\times10^{-2}$ & $\boldsymbol{9.821\times10^{0}}$ \\ 
 & 6 & $8.998\times10^{0}$ & NA & $2.333\times10^{-1}$ & $1.239\times10^{-2}$ & NA & $7.809\times10^{-3}$ & $3.652\times10^{-2}$ & $\boldsymbol{1.075\times10^{1}}$ \\ 
\multirow{5}{*}{\shortstack[c]{Kwon\\Index}} & 2 & $7.347\times10^{4}$ & $1.188\times10^{9}$ & $3.756\times10^{5}$ & $3.842\times10^{5}$ & $\boldsymbol{1.243\times10^{4}}$ & $6.039\times10^{5}$ & $8.183\times10^{4}$ & $1.513\times10^{4}$ \\ 
 & 3 & $2.930\times10^{5}$ & $8.962\times10^{13}$ & $9.483\times10^{5}$ & $1.209\times10^{6}$ & $\boldsymbol{7.266\times10^{3}}$ & $1.523\times10^{6}$ & $1.560\times10^{6}$ & $3.819\times10^{4}$ \\ 
 & 4 & $2.974\times10^{6}$ & $1.956\times10^{13}$ & $1.949\times10^{6}$ & $1.767\times10^{6}$ & NA & $3.817\times10^{6}$ & $1.852\times10^{6}$ & $\boldsymbol{5.646\times10^{5}}$ \\ 
 & 5 & $2.508\times10^{6}$ & $7.710\times10^{14}$ & $1.119\times10^{10}$ & $2.802\times10^{6}$ & NA & $5.217\times10^{6}$ & $1.032\times10^{7}$ & $\boldsymbol{1.179\times10^{6}}$ \\ 
 & 6 & $2.453\times10^{6}$ & NA & $2.818\times10^{6}$ & $\boldsymbol{2.320\times10^{6}}$ & NA & $4.825\times10^{6}$ & $7.025\times10^{6}$ & $4.574\times10^{6}$ \\
   \bottomrule
\end{tabular}
}
\end{table}

\begin{table}[ht]
\centering
\caption{Computational cost summary for Case 3} 
\label{tab:eval3c}
\resizebox{\textwidth}{!}{
\begin{tabular}{M{0.1\textwidth}rllllllll}
  \toprule
Criteria & c & ABC & CLASSIC & FPA & GSA & HHO & IFA & PSO & TLBO \\ 
  \midrule
\multirow{5}{*}{\shortstack[c]{Itera-\\tion}} & 2 & $2.374\times10^{1}$ & $1.824\times10^{1}$ & $\boldsymbol{1.058\times10^{1}}$ & $1.078\times10^{1}$ & $1.382\times10^{1}$ & $1.272\times10^{1}$ & $1.832\times10^{1}$ & $2.267\times10^{1}$ \\ 
 & 3 & $3.295\times10^{1}$ & $2.121\times10^{1}$ & $\boldsymbol{1.061\times10^{1}}$ & $1.130\times10^{1}$ & $1.850\times10^{1}$ & $1.276\times10^{1}$ & $1.816\times10^{1}$ & $2.536\times10^{1}$ \\ 
 & 4 & $3.403\times10^{1}$ & $2.500\times10^{1}$ & $1.072\times10^{1}$ & $\boldsymbol{1.060\times10^{1}}$ & NA & $1.276\times10^{1}$ & $1.791\times10^{1}$ & $2.938\times10^{1}$ \\ 
 & 5 & $4.175\times10^{1}$ & $2.867\times10^{1}$ & $\boldsymbol{1.073\times10^{1}}$ & $1.132\times10^{1}$ & NA & $1.267\times10^{1}$ & $1.811\times10^{1}$ & $2.973\times10^{1}$ \\ 
 & 6 & $3.769\times10^{1}$ & NA & $1.101\times10^{1}$ & $\boldsymbol{1.054\times10^{1}}$ & NA & $1.263\times10^{1}$ & $1.799\times10^{1}$ & $3.314\times10^{1}$ \\ 
\multirow{5}{*}{\shortstack[c]{Comp.\\time}} & 2 & $5.790\times10^{1}$ & $\boldsymbol{9.467\times10^{-1}}$ & $7.336\times10^{0}$ & $7.189\times10^{0}$ & $8.277\times10^{0}$ & $2.134\times10^{1}$ & $1.096\times10^{1}$ & $3.975\times10^{1}$ \\ 
 & 3 & $8.271\times10^{1}$ & $\boldsymbol{1.155\times10^{0}}$ & $7.060\times10^{0}$ & $7.612\times10^{0}$ & $1.229\times10^{1}$ & $2.071\times10^{1}$ & $1.074\times10^{1}$ & $3.902\times10^{1}$ \\ 
 & 4 & $8.855\times10^{1}$ & $\boldsymbol{1.390\times10^{0}}$ & $7.472\times10^{0}$ & $7.718\times10^{0}$ & NA & $2.138\times10^{1}$ & $1.506\times10^{1}$ & $4.015\times10^{1}$ \\ 
 & 5 & $1.122\times10^{2}$ & $\boldsymbol{1.843\times10^{0}}$ & $7.387\times10^{0}$ & $7.736\times10^{0}$ & NA & $2.379\times10^{1}$ & $9.287\times10^{0}$ & $3.591\times10^{1}$ \\ 
 & 6 & $1.059\times10^{2}$ & NA & $8.267\times10^{0}$ & $\boldsymbol{7.591\times10^{0}}$ & NA & $2.428\times10^{1}$ & $9.586\times10^{0}$ & $4.155\times10^{1}$ \\ 
   \bottomrule
\end{tabular}
}
\end{table}

\subsection{Statistical analysis}

To confirm the significance of our analysis results, we also performed a statistical analysis to investigate whether there is a significant difference in the evaluation measurements between the algorithms in each case. We quantify this using Kruskal-Wallis as an alternative to ANOVA which requires normality and homogeneity of variance assumption. 

\begin{figure}[htbp]
    \centering
    \includegraphics[width=\textwidth]{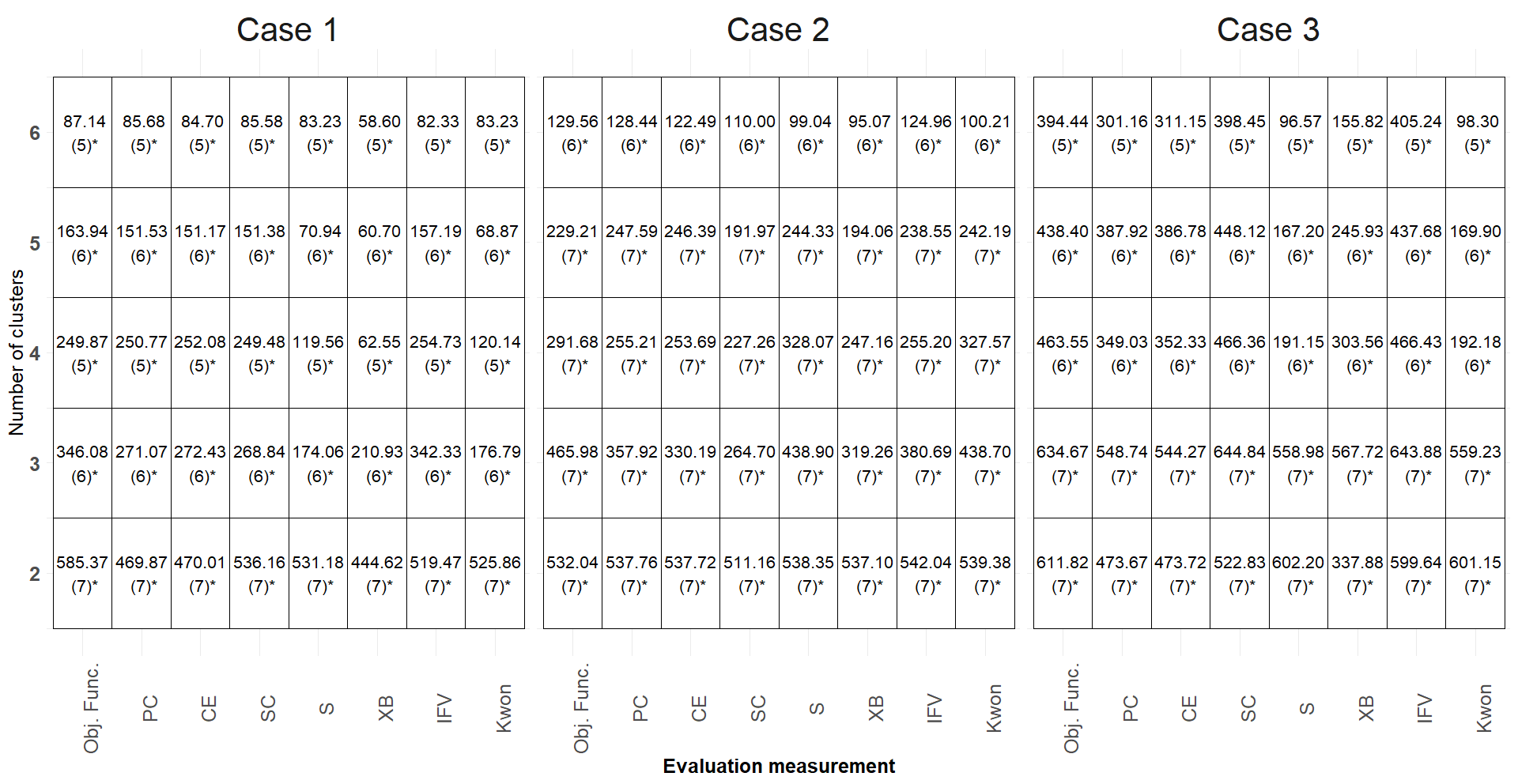}
    \caption{Kruskal-Wallis results of the evaluation measurements difference between FGWC algorithms based on the number of cases and number of clusters. The values in the grid represents Kruskal-Wallis statistics, while in bracket is the degrees of freedom. The asterisk (*) sign denotes significant difference of values between algorithms.}
    \label{fig:kruskal}
\end{figure}

The results can be seen in Figure~\ref{fig:kruskal}. From the figure, it can be seen that the Kruskal-Wallis statistics produced is relatively high. On the other hand, the degrees of freedom are different because some algorithms, such as the classical and HHO algorithms, did not produce proper results. Therefore, given the statistics and degrees of freedom, we can conclude that there is a significant difference between the FGWC algorithms in each case, confiming that the choice of optimisation algorithms in each case becomes important.

\subsection{Overview of optimised FGWC spatial clustering results}

\begin{figure}[htbp]
    \centering
    \begin{subfigure}[b]{0.45\textwidth}
        \centering
        \includegraphics[width=\textwidth]{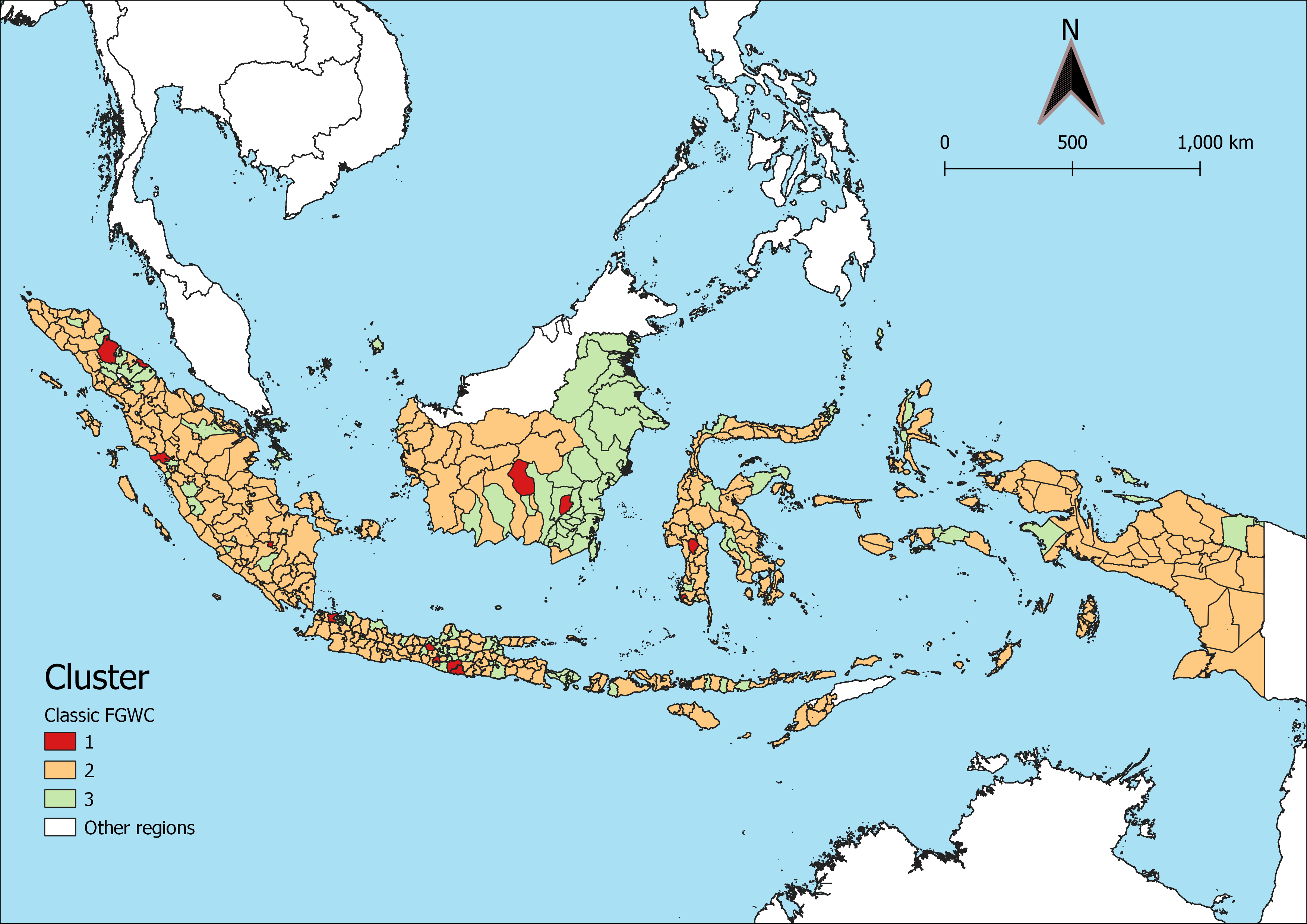}
        \caption{Classic FGWC}
        \label{subfig:map_fgwc}
    \end{subfigure}
    \hfill
    \begin{subfigure}[b]{0.45\textwidth}
        \centering
        \includegraphics[width=\textwidth]{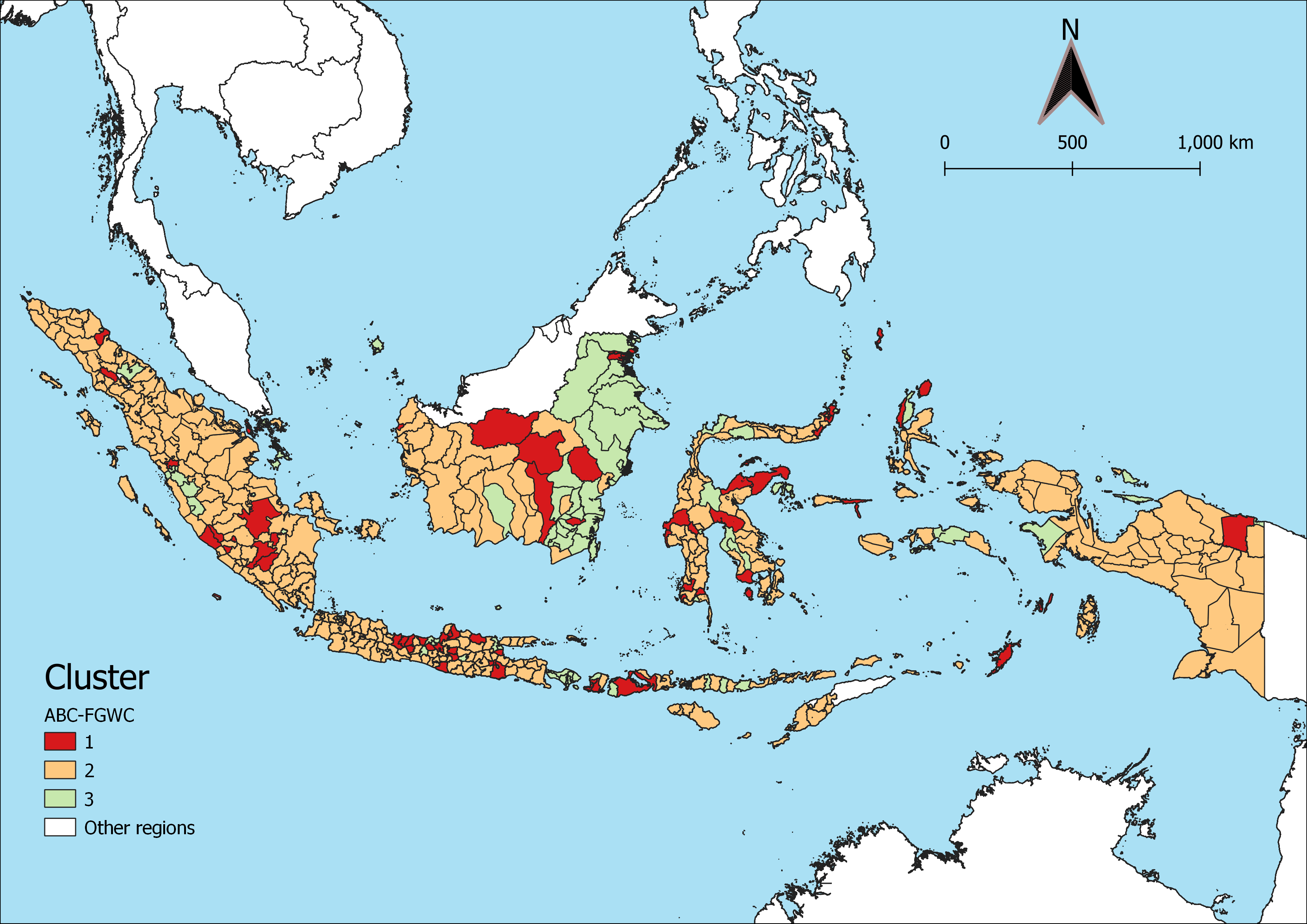}
        \caption{Artificial Bee Colony}
        \label{subfig:map_abc}
    \end{subfigure}
    \hfill
    \begin{subfigure}[b]{0.45\textwidth}
        \centering
        \includegraphics[width=\textwidth]{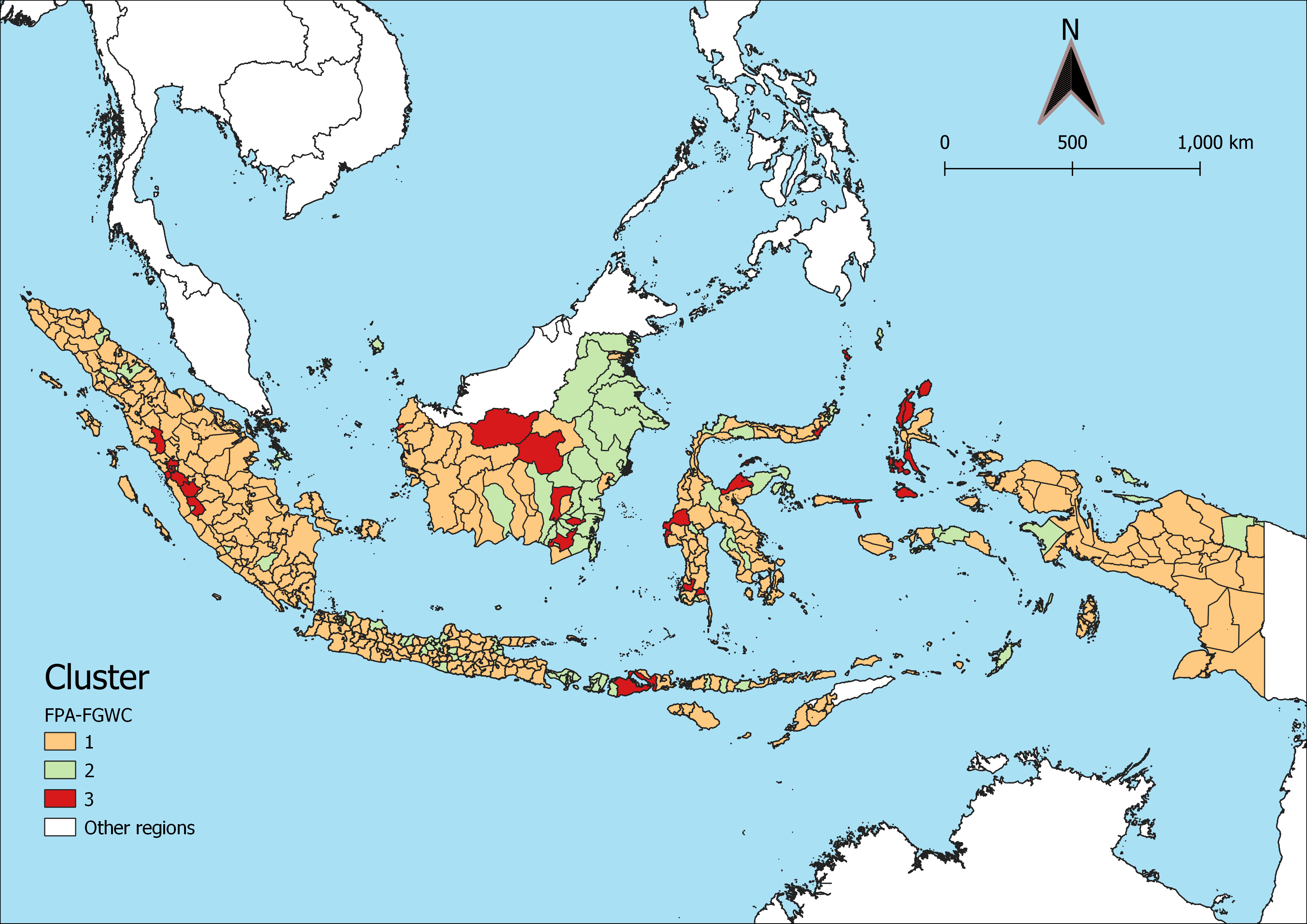}
        \caption{Flower Pollination Algorithm}
        \label{subfig:map_fpa}
    \end{subfigure}
    \hfill
    \begin{subfigure}[b]{0.45\textwidth}
        \centering
        \includegraphics[width=\textwidth]{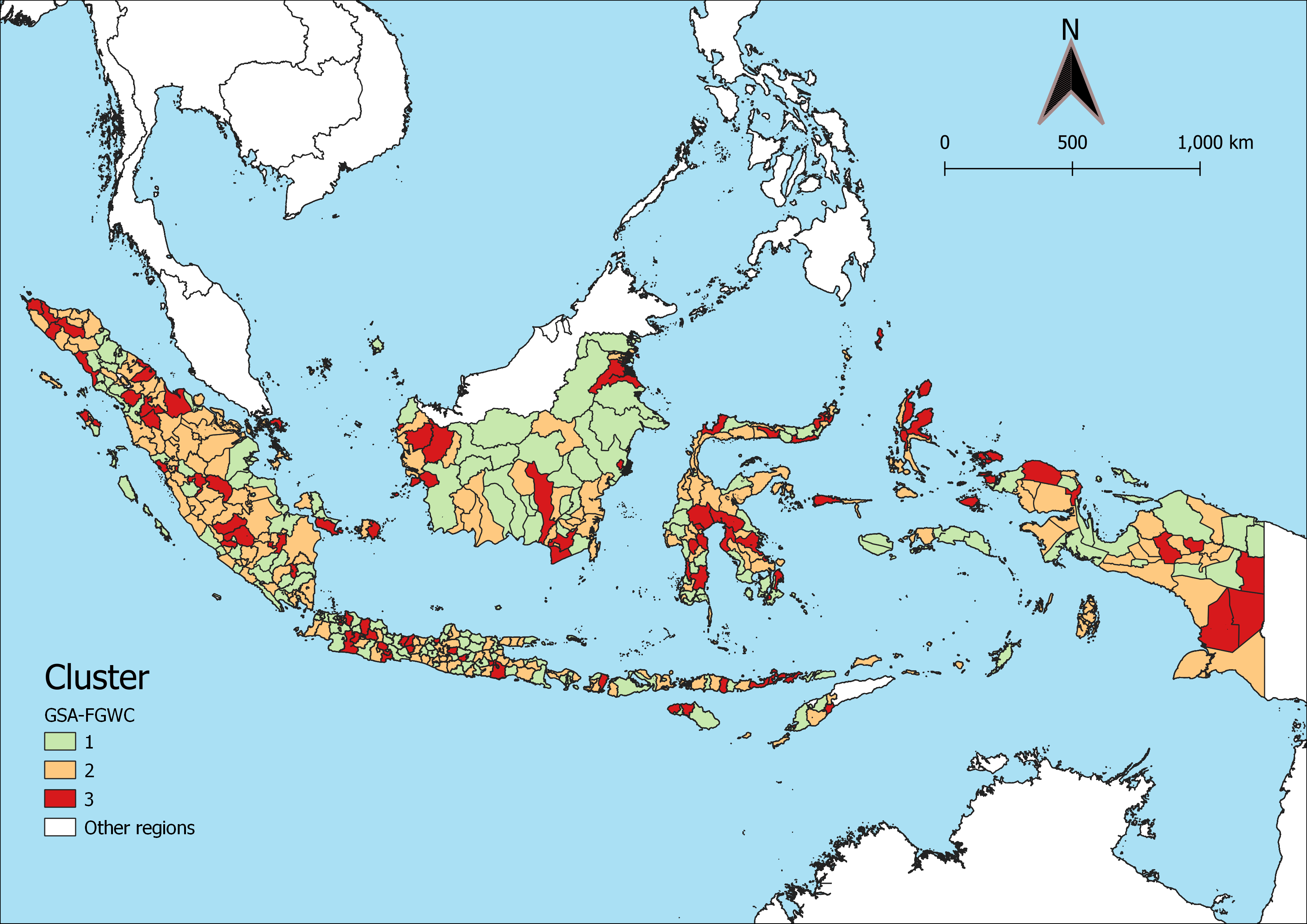}
        \caption{Gravitational Search Algorithm}
        \label{subfig:map_gsa}
    \end{subfigure}
    \caption{Spatial map on social vulnerability (case 3) clustering in Indonesia at district level using 3 clusters based on optimisation algorithms}
    \label{fig:map1}
\end{figure}

\begin{figure}[htbp]
    \centering
    \begin{subfigure}[b]{0.45\textwidth}
        \centering
        \includegraphics[width=\textwidth]{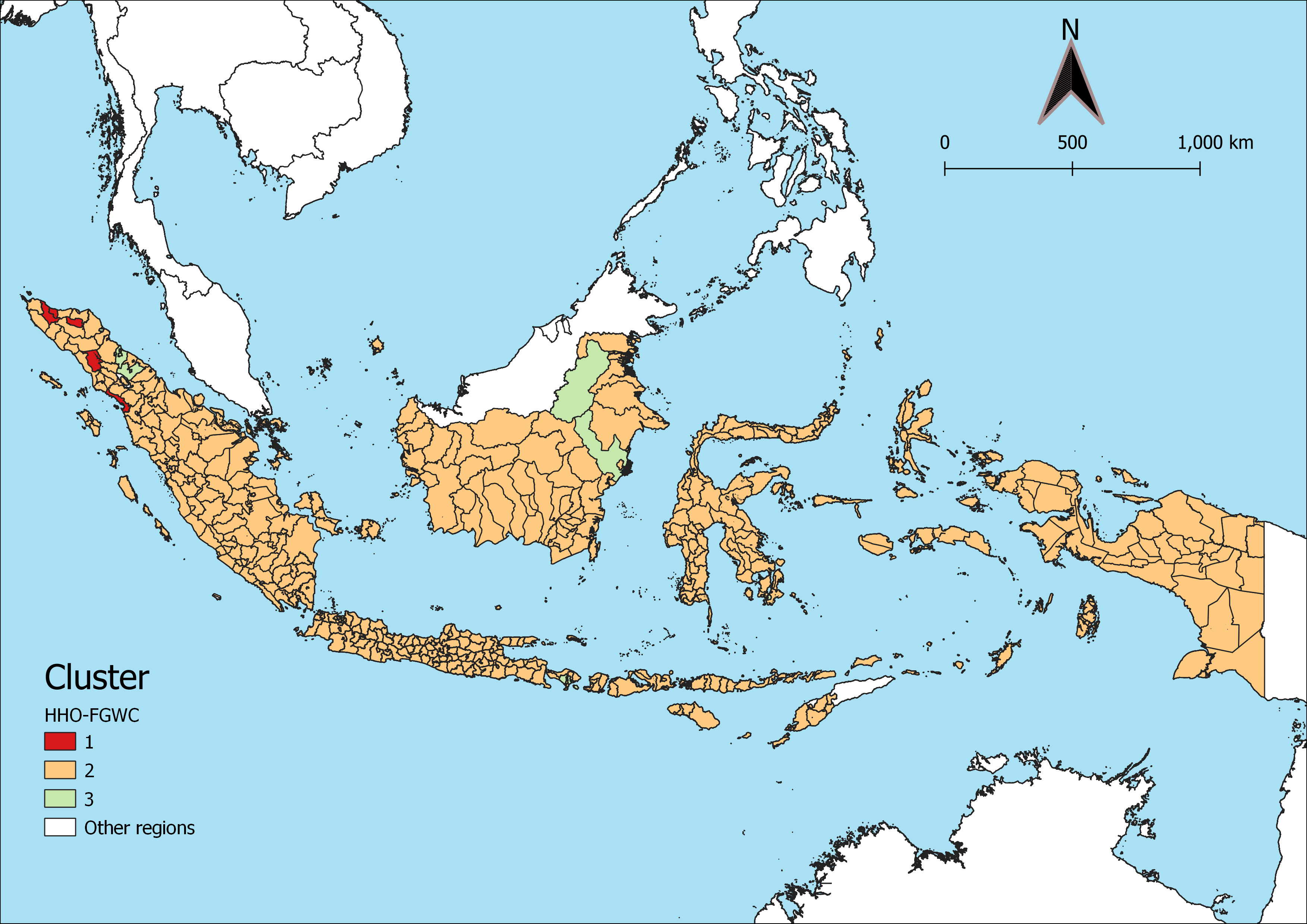}
        \caption{Harris-Hawk Optimisation}
        \label{subfig:map_hho}
    \end{subfigure}
    \hfill
    \begin{subfigure}[b]{0.45\textwidth}
        \centering
        \includegraphics[width=\textwidth]{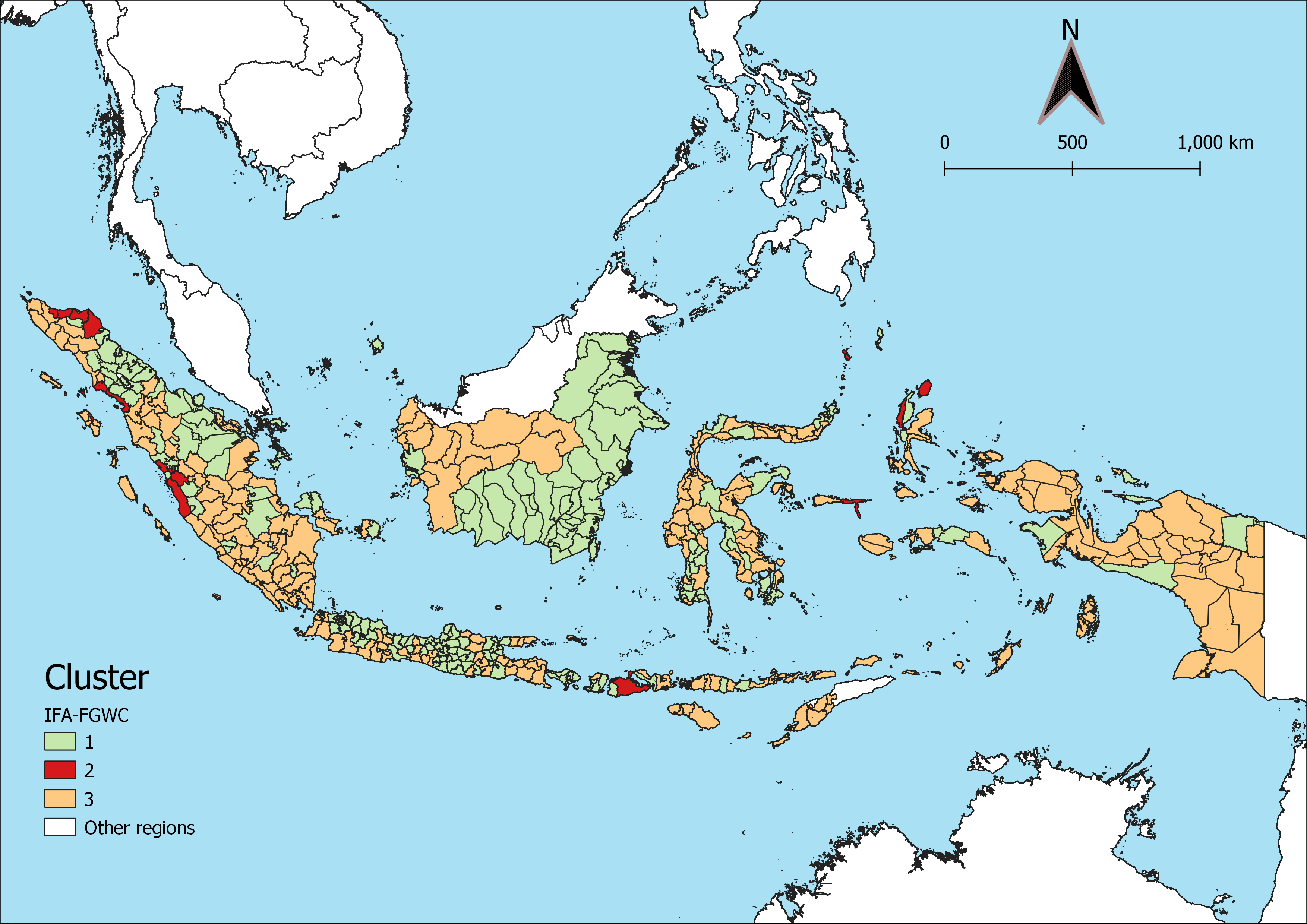}
        \caption{Intelligent Firefly Algorithm}
        \label{subfig:map_ifa}
    \end{subfigure}
    \hfill
    \begin{subfigure}[b]{0.45\textwidth}
        \centering
        \includegraphics[width=\textwidth]{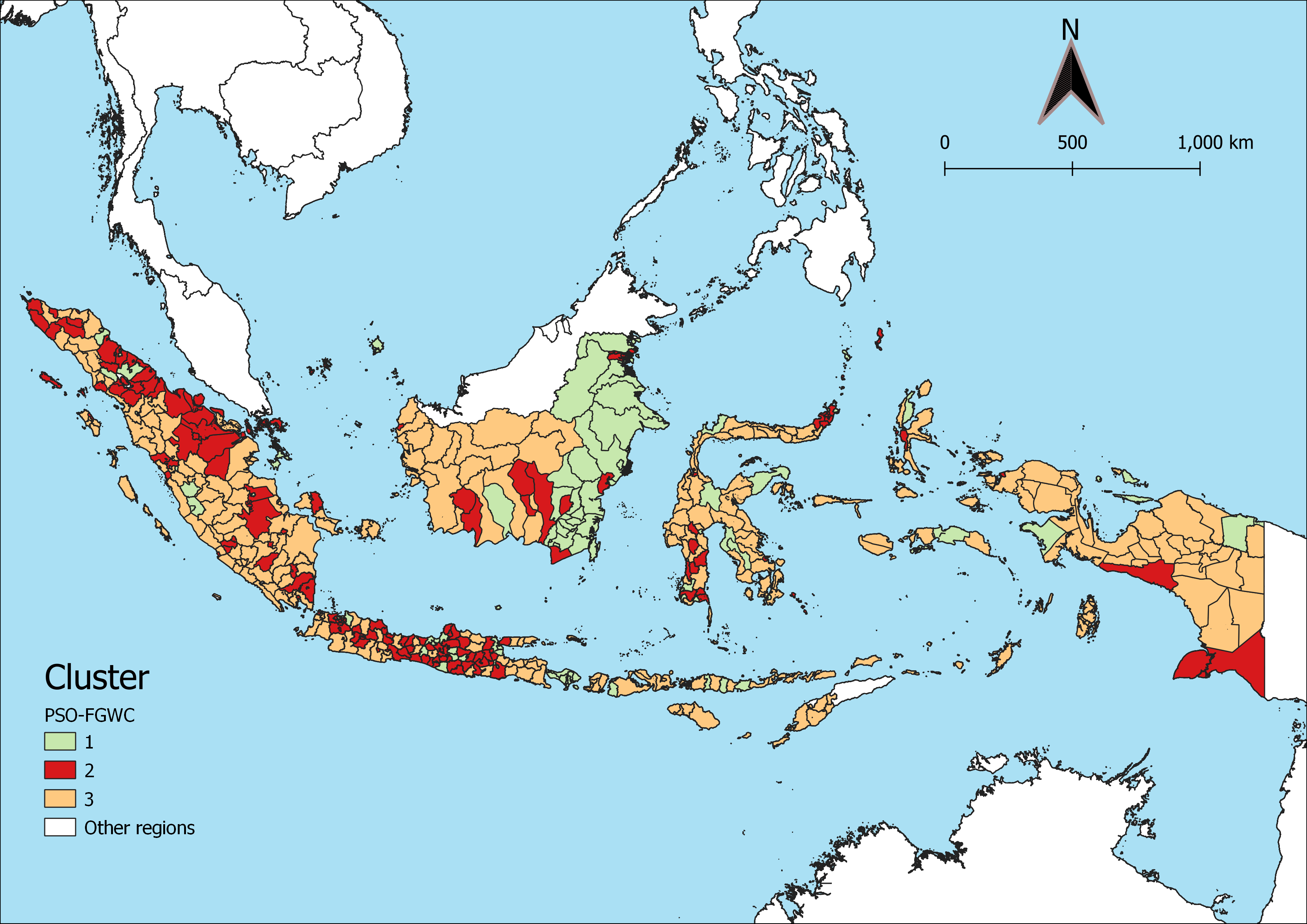}
        \caption{Particle Swarm Optimisation}
        \label{subfig:map_pso}
    \end{subfigure}
    \hfill
    \begin{subfigure}[b]{0.45\textwidth}
        \centering
        \includegraphics[width=\textwidth]{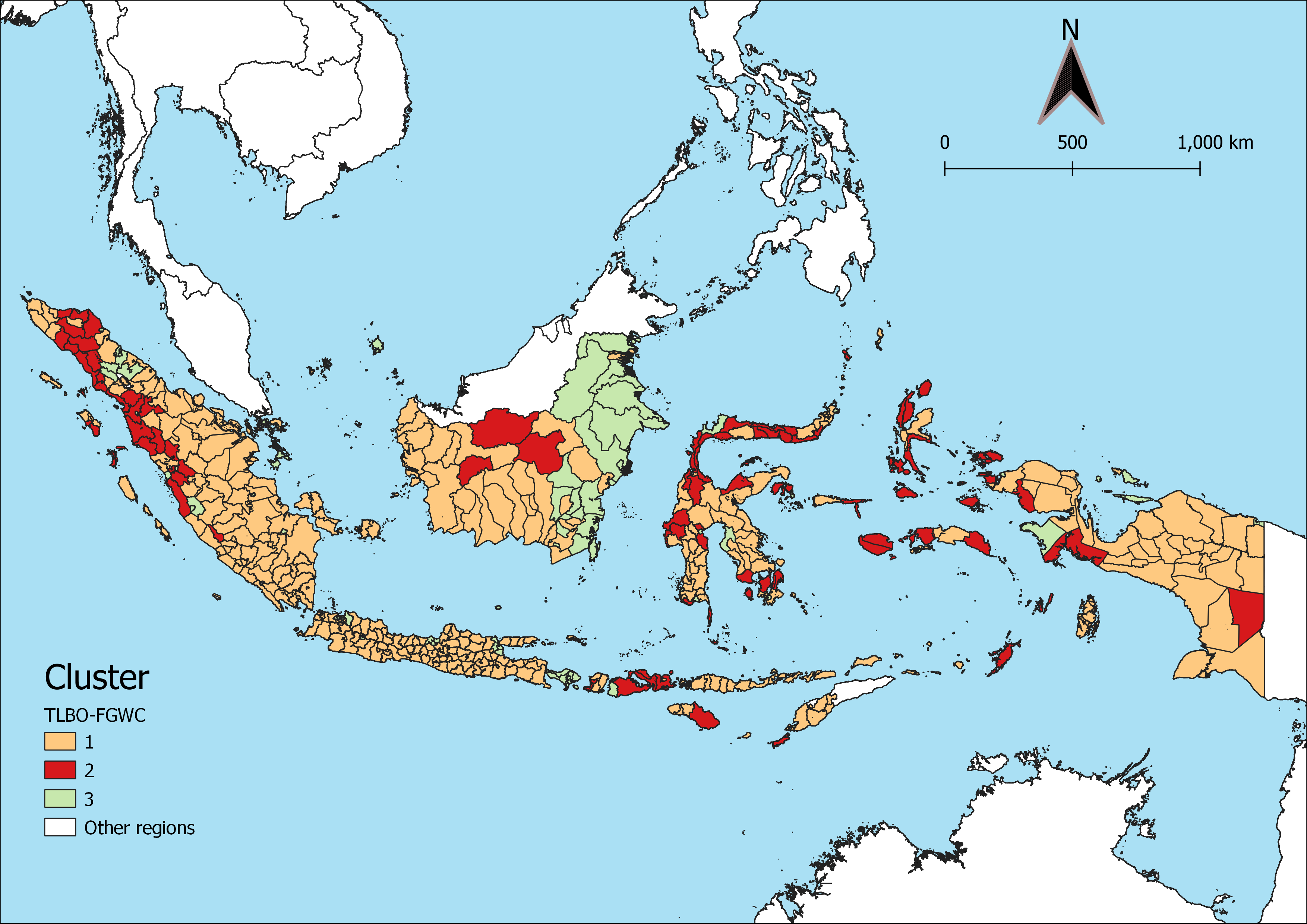}
        \caption{Teaching-Learning Based Optimisation}
        \label{subfig:map_tlbo}
    \end{subfigure}
    \caption{Spatial map on social vulnerability (case 3) clustering in Indonesia at district level using 3 clusters based on optimisation algorithms (Continued)}
    \label{fig:map}
\end{figure}

In this subsection, we provide an overview and critical analysis of the FGWC clustering results. Given the demonstrated potential of most algorithms to optimise FGWC, particularly in Case 3, we illustrate clustering results using an example with three clusters for each optimisation algorithm. Figures~\ref{fig:map} and~\ref{fig:map1} depict spatial cluster formations resulting from different optimisation methods.

Generally, maps reveal clear spatial patterns. Districts that are geographically close frequently belong to the same clusters. For example, regions in East and North Kalimantan predominantly fall into a single cluster, although some districts deviate from this trend. These spatial patterns are likely the result of incorporating spatial influence into the membership calculation during the fuzzy clustering process.

However, notable differences arise between the clustering results when comparing algorithms. The classic FGWC, FPA and HHO (Figures~\ref{subfig:map_fgwc},\ref{subfig:map_fpa}, and\ref{subfig:map_hho}) tend to generate predominantly large yellow clusters that cover most of Indonesia, accompanied by smaller red and green clusters. In contrast, ABC and TLBO (Figures~\ref{subfig:map_abc} and~\ref{subfig:map_tlbo}) result in clearly delineated red clusters that appear consistently across adjacent areas in Sumatera, Kalimantan, and eastern Indonesia. Similar spatially coherent patterns also appear in the results of IFA and PSO (Figures~\ref{subfig:map_ifa} and~\ref{subfig:map_pso}), where the clusters clearly group neighboring areas predominantly in red or green. 

Therefore, it can be concluded that the different algorithms produce varying clustering results, emphasising the need to carefully select an appropriate algorithm tailored to specific research objectives and data characteristics. These variations comes primarily from the differing optimisation strategies each algorithm employs, influencing their capability to explore diverse spatial patterns and manage initial condition sensitivities. The findings also encourage policymakers and researchers to investigate algorithm performance considering both clustering evaluations and spatial coherence to ensure robust and actionable outcomes.

\section{Discussions} 
\label{sec:disc}


\subsection{Reflections on previous studies}

Our findings demonstrate that optimisation algorithms generally outperform the classic FGWC method in terms of cluster evaluation and expected cluster formations. This aligns with previous research confirming the benefits of optimisation techniques in improving fuzzy clustering results, particularly when spatial effects are considered~\citep{fgwcabc1,fgwcpso,carakafpa,fgwcgsa}. Specifically, TLBO emerged as the most effective algorithm for optimising FGWC clustering quality, supporting the findings of~\cite{Nayak2018} and~\cite{nasutiontlbo}, who identified TLBO as superior for overcoming the limitations of FCM and FGWC compared to other optimisation algorithms. Furthermore, ABC exhibited great potential to optimise FGWC, corroborating earlier findings by~\citep{fgwcabc1,fgwcabc2}. However, despite their enhanced clustering performance, these algorithms may occasionally fail to form the expected clusters due to dependencies among membership assignments, resulting in unintended outcomes~\citep{grover2014,Thomas2012ComparativeAO}. Consequently, although the ABC and TLBO algorithms provided by the \texttt{naspaclust} package significantly improve cluster quality, further development and refinement of the underlying fuzzy clustering techniques remain essential.

We also found that the optimisation algorithms in FGWC can produce a spatial pattern with different results each other. These differences align with previous research by~\cite{Nasution2020,NASUTIONfpa} and~\cite{carakafpa}, which shows that different FGWC optimisation algorithms can create compact spatial clusters due to their geographical weighting and optimisation feature. Each algorithm identifies different optimum points, resulting in diverse membership patterns and, consequently, varying spatial cluster formations. Such variations arise from distinctive search mechanisms and the balance between exploration and exploitation strategies inherent in each algorithm. Algorithms such as ABC and TLBO effectively explore broader solution spaces, resulting in meaningful and coherent spatial clusters, such as in~\cite{fgwcabc2} in ABC and~\cite{nasution2018} in TLBO. In contrast, algorithms that emphasise rapid convergence, such as HHO, may settle prematurely into suboptimal solutions, producing less distinct spatial clusters. This problem also justified the choice of using rapid progressive dives in HHO when clustering the green infrastructure vulnerability in Indonesia, as done by~\cite{Sumargo2024}. 

Moreover, this observation is consistent with the "no free lunch" theorem, which suggests that no single algorithm universally solves all problems optimally~\citep{yang2014,nasution2018}. Each optimisation algorithm offers distinct competitive advantages; for example, TLBO excels at avoiding local optima at the cost of higher computational demand, while other algorithms may rapidly converge but risk local optima traps. Consequently, selecting an algorithm from \texttt{naspaclust} depends on the user's priorities: whether efficiency or clustering quality is paramount. Furthermore, to mitigate the risks of unintended cluster formation, careful initialisation is recommended using parameters such as \texttt{randomN} in FGWC is recommended.

On the other hand, the practical implications of these findings for policymaking are significant. Policymakers must carefully select algorithms aligning with their specific analytical objectives, considering both clustering accuracy and spatial coherence. Algorithms such as TLBO or ABC, which demonstrate robust spatial performance, are preferable choices. In addition, policymakers should consider clustering methods in groups or validate cluster solutions using expert judgment to enhance reliability and ensure robust policy applications from clustering spatial analyses.

\subsection{Limitations and Future Directions}

In addition to the introductions of this package, this study still has some limitations. First, the experiment in this study is only limited to the default features of the optimisation algorithms because the goal is to introduce the \texttt{naspaclust} package in R. Furthermore, we also only use benchmark data that have been used in previous studies using FGWC as the algorithms. 

Thus, the limitations open many possibilities to continue the optimisation algorithms in FGWC studies. Firstly, one may apply these algorithms to solve real-world problems that require the clustering algorithm at the regional level using different optimised FGWC. Using spatial information may give more insight into the clustering results. Considering the features available in the \texttt{naspaclust} package, one can implement the enhancement version of optimisation algorithms such as in GSA, HHO and TLBO based on Section~\ref{sec:back}. Furthermore, the modification can also be done based on distance using the available distance features in the \texttt{ rdist} package~\citep{rdist}. Other modifications that can be made are from randomisation terms such as in IFA, FPA, and GSA, or the update inertia weight in PSO. Considering more optimal results in Cases 2 and 3, we recommend using dimensionality reduction techniques such as PCA and t-SNE for data with high dimensions before doing clustering.

\section{Conclusions and Future Studies}
\label{sec:conc}
This study aims to introduce the \texttt{naspaclust} package which utilises the nature-inspired optimisation algorithms to improve fuzzy clustering performance, particularly FGWC. We have briefly described the implementation of the package in R software as well as providing the performance evaluation of some optimisation algorithms. This study found that the optimisation algorithms is proven to improve the performance of FGWC in any type of data and produce spatial patterns of the clustering results. Moreover, TLBO has the most potential to optimise FGWC in all cases. However, the available algorithms still have potential disadvantages in finding unexpected cluster formation which is not consistent with the initialised number of clusters in certain simulations. Therefore, we hope that this paper could become foundation for everyone wants to do GDA which needs spatial clustering techniques to improve the cluster quality.


\section*{Conflict of Interest}
The authors declare that there are no conflicts of interest with respect to the publication of this article. The research and its results were conducted without financial, personal, or professional relationships that could potentially influence or bias the findings and interpretations presented in this work.

\section*{Data availability statement }
The data that support the simulations of this study are available in the GitHub repository of \texttt{naspaclust} 
\href{https://github.com/bmlmcmc/naspaclust/tree/main/data}{here}
.

\bibliographystyle{elsarticle-harv} 
\bibliography{cas-refs}

@article{ZADEH1965338,
title = {Fuzzy sets},
journal = {Information and Control},
volume = {8},
number = {3},
pages = {338-353},
year = {1965},
issn = {0019-9958},
doi = {https://doi.org/10.1016/S0019-9958(65)90241-X},
url = {https://www.sciencedirect.com/science/article/pii/S001999586590241X},
author = {L.A. Zadeh}
}

@inproceedings{fgwcabc1,
author={Wijayanto, Arie Wahyu and Purwarianti, Ayu},
title={Improvement design of fuzzy geo-demographic clustering using Artificial Bee Colony optimization},
booktitle={2014 International Conference on Cyber and IT Service Management (CITSM)},
year={2014},
digitaldoi={10.1109/CITSM.2014.7042178},
isbn={978-1-4799-7975-2},
month={nov},
number={},
pages={69--74},
publisher={IEEE},
}

@article{fgwcabc2,
author={Wijayanto, Arie Wahyu and Purwarianti, Ayu and Son, Le Hoang},
title={Fuzzy geographically weighted clustering using artificial bee colony: An efficient geo-demographic analysis algorithm and applications to the analysis of crime behavior in population},
journal={Applied Intelligence},
year={2016},
digitaldoi={10.1007/s10489-015-0705-7},
issn={0924-669X},
month={mar},
number={2},
pages={377--398},
volume={44},
publisher={Springer},
}

@inproceedings{fgwcpso,
author={Wijayanto, Arie Wahyu and Purwarianti, Ayu},
title={Improvement of fuzzy geographically weighted clustering using particle swarm optimization},
booktitle={2014 International Conference on Information Technology Systems and Innovation (ICITSI)},
year={2014},
digitaldoi={10.1109/ICITSI.2014.7048229},
isbn={978-1-4799-6527-4},
month={November},
pages={7--12},
publisher={IEEE},
}

@inproceedings{fgwcaco,
author={Wijayanto, Arie Wahyu and Mariyah, Siti and Purwarianti, Ayu},
title={Enhancing clustering quality of fuzzy geographically weighted clustering using Ant Colony optimization},
booktitle={2017 International Conference on Data and Software Engineering (ICoDSE)},
year={2017},
digitaldoi={10.1109/ICODSE.2017.8285858},
isbn={978-1-5386-1449-5},
month={nov},
pages={1--6},
publisher={IEEE},
}

@inproceedings{fgwc, 
author={G. A. Mason and R. D. Jacobson}, 
title={Fuzzy Geographically Weighted Clustering}, 
booktitle={Proceedings of the 9th International Conference on Geocomputation}, 
year={2007}, 
pages={1-7}, 
number={},
}

@inproceedings{fengflowerdew, 
author={Z. Q. Feng and R. Flowerdew}, 
title={Fuzzy geodemographics: a contribution from fuzzy clustering methods}, 
booktitle={Innovations in GIS 5: Selected Papers from the Fifth National Conference on GIS Research UK (GISRUK)}, 
year={1998}, 
pages={112–120}, 
publisher={Taylor and Francis Group},
}

@article{paez,
author={Antonio Paez and Martin Trepanier and Catherine Morency},
title={Geodemographic analysis and the identification of potential business partnerships enabled by transit smart cards },
journal={Transportation Research Part A: Policy and Practice },
volume={45},
number={7},
pages={640-652},
year={2011},
note={},
issn={0965-8564},
digitaldoi={10.1016/j.tra.2011.04.002},
}

@article{kwon,
author={Kwon, S. H.},
title={Cluster validity index and fuzzy clustering},
journal={Electronics Letters},
ISSN={},
publisher={},
year={1998},
month={},
volume={34},
number={},
pages={2176-2177},
}

@inproceedings{nasution2018, 
author={B. I. Nasution and R. Kurniawan}, 
booktitle={2018 International Conference on Information and Communications Technology (ICOIACT)}, 
title={Robustness of classical fuzzy C-means (FCM)}, 
year={2018}, 
volume={}, 
number={}, 
pages={321-325}, 
digitaldoi={10.1109/ICOIACT.2018.8350729}, 
ISSN={}, 
month={March},
}

@article{Singh2020a,
author = {Tribhuvan Singh},
doi = {10.1007/s00521-020-04951-2},
issn = {1433-3058},
journal = {Neural Computing and Applications},
number = {23},
pages = {17789--17803},
title = {{A chaotic sequence-guided Harris hawks optimizer for data clustering}},
url = {https://doi.org/10.1007/s00521-020-04951-2},
volume = {32},
year = {2020}
}

@article{AroraAnand2017,
  place = {NL},
  title = {Chaos-enhanced flower pollination algorithms for global optimization},
  volume = {33},
  ISSN = {10641246,  18758967},
  url = {https://doi.org/10.3233/JIFS-17708},
  DOI = {10.3233/JIFS-17708},
  abstractNote = {The flower pollination algorithm (FPA) is a recently developed meta-heuristic algorithm inspired by the pollination process of flowers. Similar to other meta-heuristic algorithms,  it encounters two probable problems,  i.e.,  entrapment in local optima and slow convergence speed,  in solving challenging complex real world problems. Similar to the chaos in actual flower pollination process,  this paper proposes new FPAs that employ chaotic maps for adjustment of parameters with the aim to improve the convergence rate and prevent the FPAs to get trapped on local optima. This is achieved by employing chaotic number generators every time,  a random number is needed by the classical FPA. Two new chaotic FPAs have been proposed and various test problems are used for their performance evaluation. To check the effectiveness of the proposed algorithms,  they are tested on various benchmark functions and engineering design problems with different characteristics having real world applications. The simulation results demonstrate that the chaotic maps are able to significantly boost the performance of FPAs.},
  number = {6},
  journal = {Journal of Intelligent \& Fuzzy Systems},
  publisher = {IOS Press},
  author = {Arora,  Sankalap and Anand,  Priyanka},
  year = {2017},
  month = Nov,
  pages = {3853–3869}
}

@article{Kaur2020,
  doi = {10.1016/j.asoc.2019.105523},
  url = {https://doi.org/10.1016/j.asoc.2019.105523},
  year = {2020},
  month = dec,
  publisher = {Elsevier {BV}},
  volume = {97},
  pages = {105523},
  author = {Arvinder Kaur and Saibal Kumar Pal and Amrit Pal Singh},
  title = {Hybridization of Chaos and Flower Pollination Algorithm over K-Means for data clustering},
  journal = {Applied Soft Computing}
}

@Article{carakafpa,
AUTHOR = {Caraka, Rezzy Eko and Kurniawan, Robert and Nasution, Bahrul Ilmi and Jamilatuzzahro, Jamilatuzzahro and Gio, Prana Ugiana and Basyuni, Mohammad and Pardamean, Bens},
TITLE = {Micro, Small, and Medium Enterprises’ Business Vulnerability Cluster in Indonesia: An Analysis Using Optimized Fuzzy Geodemographic Clustering},
JOURNAL = {Sustainability},
VOLUME = {13},
YEAR = {2021},
NUMBER = {14},
ARTICLE-NUMBER = {7807},
URL = {https://www.mdpi.com/2071-1050/13/14/7807},
ISSN = {2071-1050},
DOI = {10.3390/su13147807}
}

@book{bezdek1981,
author={Bezdek, James C.},
title={Pattern Recognition with Fuzzy Objective Function Algorithms},
year={1981},
isbn={0306406713},
publisher={Kluwer Academic Publishers},
address={Norwell, MA, USA},
}

@article{bezdek1984,
author={James C. Bezdek and Robert Ehrlich and William Full},
title={FCM: The fuzzy c-means clustering algorithm},
journal={Computers \& Geosciences},
volume={10},
number={2},
pages={191 - 203},
year={1984},
issn={0098-3004},
digitaldoi={https://doi.org/10.1016/0098-3004(84)90020-7},
digitalurl={http://www.sciencedirect.com/science/article/pii/0098300484900207},
}

@article{gueorguieva,
author={Natacha Gueorguieva and Iren Valova and George Georgiev},
title={M\&MFCM: Fuzzy C-means Clustering with Mahalanobis and Minkowski Distance Metrics},
journal={Procedia Computer Science},
volume={114},
pages={224 - 233},
year={2017},
note={Complex Adaptive Systems Conference with Theme: Engineering Cyber Physical Systems, CAS October 30 – November 1, 2017, Chicago, Illinois, USA},
issn={1877-0509},
digitaldoi={https://doi.org/10.1016/j.procs.2017.09.064},
digitalurl={http://www.sciencedirect.com/science/article/pii/S1877050917318689},
}

@book{han2011,
author={Han, J. and Pei, J. and Kamber, M.},
title={Data Mining: Concepts and Techniques},
isbn={9780123814807},
series={The Morgan Kaufmann Series in Data Management Systems},
year={2011},
publisher={Elsevier Science}
}

@article{izakian,
author={Izakian, Hesam and Abraham, Ajith},
title={Fuzzy C-means and Fuzzy Swarm for Fuzzy Clustering Problem},
journal={Expert Syst. Appl.},
issue_date={March, 2011},
volume={38},
number={3},
month={mar},
year={2011},
issn={0957-4174},
pages={1835--1838},
numpages={4},
digitalurl={http://dx.doi.org/10.1016/j.eswa.2010.07.112},
digitaldoi={10.1016/j.eswa.2010.07.112},
acmid={1879373},
publisher={Pergamon Press, Inc.},
address={Tarrytown, NY, USA},
}

@book{yang2014,
author={Yang, X.S.},
title={Nature-Inspired Optimization Algorithms},
isbn={9780124167452},
lccn={2014931594},
series={Elsevier insights},
digitalurl={https://books.google.co.jp/books?id=BbHVAQAAQBAJ},
year={2014},
publisher={Elsevier Science}
}

@article{wu2012, 
author={J. Wu and H. Xiong and C. Liu and J. Chen}, 
title={A Generalization of Distance Functions for Fuzzy$c$-Means Clustering With Centroids of Arithmetic Means}, 
journal={IEEE Transactions on Fuzzy Systems}, 
year={2012}, 
volume={20}, 
number={3}, 
pages={557-571}, 
digitaldoi={10.1109/TFUZZ.2011.2179659}, 
ISSN={1063-6706}, 
month={June},
}

@article{singh2013,
author={Archana Singh and Avantika Yadav and Ajay Rana},
title={K-means with Three different Distance Metrics},
journal={International Journal of Computer Applications},
year={2013},
volume={67},
number={10},
pages={13-17},
month={April},
}

@inproceedings{srivastava2011,
author={Srivastava, Vivek and Tripathi, Bipin K. and Pathak, Vinay K.},
title={An Evolutionary Fuzzy Clustering with Minkowski Distances},
editor={Lu, Bao-Liang and Zhang, Liqing and Kwok, James},
booktitle={Neural Information Processing},
year={2011},
publisher={Springer Berlin Heidelberg},
address={Berlin, Heidelberg},
pages={753--760},
isbn={978-3-642-24958-7},
}

@inproceedings{liu2008, 
author={Hsiang-Chuan Liu and Jeng-Ming Yih and Der-Bang Wu and Shin-Wu Liu}, 
title={Fuzzy C-mean algorithm based on “complete” Mahalanobis distances}, 
booktitle={2008 International Conference on Machine Learning and Cybernetics}, 
year={2008}, 
volume={6}, 
number={}, 
pages={3569-3574},
digitaldoi={10.1109/ICMLC.2008.4621023}, 
ISSN={2160-133X}, 
month={July},
}

@article{grover2014,
author = {Grover, Nidhi},
year = {2014},
month = {03},
pages = {177-181},
title = {A study of various Fuzzy Clustering Algorithms},
volume = {3},
journal = {International Journal of Engineering Research},
doi = {10.17950/ijer/v3s3/310}
}

@article{Thomas2012ComparativeAO,
  title={Comparative Analysis Of Fuzzy Clustering Algorithms In Data Mining},
  author={Binsy Thomas and M. Nashipudimath},
  journal={International Journal of Advanced Research in Computer Science and Electronics Engineering},
  year={2012},
  volume={1}
}

@inproceedings{nayak,
author={Nayak, Janmenjoy and Nanda, Matrupallab and Nayak, Kamlesh and Naik, Bighnaraj and Behera, Himansu Sekhar},
title={An Improved Firefly Fuzzy C-Means (FAFCM) Algorithm for Clustering Real World Data Sets},
editor={Kumar Kundu, Malay and Mohapatra, Durga Prasad and Konar, Amit and Chakraborty, Aruna},
booktitle={Advanced Computing, Networking and Informatics- Volume 1},
year={2014},
publisher={Springer International Publishing},
address={Cham},
pages={339--348},
isbn={978-3-319-07353-8},
}

@article{Nayak2018,
  doi = {10.1016/j.asej.2016.01.010},
  url = {https://doi.org/10.1016/j.asej.2016.01.010},
  year = {2018},
  month = sep,
  publisher = {Elsevier {BV}},
  volume = {9},
  number = {3},
  pages = {379--393},
  author = {Janmenjoy Nayak and Bighnaraj Naik and D.P. Kanungo and H.S. Behera},
  title = {A hybrid elicit teaching learning based optimization with fuzzy c-means ({ETLBO}-{FCM}) algorithm for data clustering},
  journal = {Ain Shams Engineering Journal}
}

@article{dunn,
author={ J. C. Dunn },
title={A Fuzzy Relative of the ISODATA Process and Its Use in Detecting Compact Well-Separated Clusters},
journal={Journal of Cybernetics},
volume={3},
number={3},
pages={32-57},
year ={1973},
publisher={Taylor \& Francis},
digitaldoi={10.1080/01969727308546046},
digitalurl={https://doi.org/10.1080/01969727308546046},
eprint={https://doi.org/10.1080/01969727308546046},
}

@article{putra2017,
author={Fikri Handoko Putra and Robert Kurniawan},
title={Clustering for Disaster Areas Endemic Dengue Hemorrhagic Fever Based on Factors had Caused in East Java Using Fuzzy Geographically Weighted Clustering – Particle Swarm Optimization},
journal={Jurnal Aplikasi Statistika \& Komputasi Statistik},
volume={8},
number={01},
year={2017},
issn={2615-1367},	
pages={27},	
digitalurl={https://jurnal.stis.ac.id/index.php/jurnalasks/article/view/10},
}

@misc{toolbox,
author={B. Balasko and J. Abonyi and B. Feil}, 
title={Fuzzy Clustering and Data Analysis Toolbox: For Use with Matlab}, 
address={Veszprem, Hungary}, 
year={2005}
}

@Manual{spatialclust,
    title = {spatialClust: Spatial Clustering using Fuzzy Geographically Weighted Clustering},
    author = {Imam Habib Pamungkas and Setia Pramana},
    year = {2016},
    note = {R package version 1.1.1},
    url = {https://CRAN.R-project.org/package=spatialClust},
  }

@article{meteo,
author={Y. Lu and T. Ma and C. Yin and X. Xie and W. Tian and S. Zhong},
title={Implementation of the Fuzzy C-Means Clustering Algorithm in Meteorological Data}, 
journal={Int. J. Database Theory Appl.}, 
volume={6}, 
number={6}, 
page={1–18}, 
month={Dec},
year={2013},
}

@article{ifv,
author={K.S. Chuang and H.L. Tzeng and S. Chen and J. Wu and T.J. Chen}, 
title={Fuzzy c-means clustering with spatial information for image segmentation},
journal={Comput. Med. Imaging Graph.}, 
volume={30}, 
number={1}, 
page={9–15}, 
month={Jan}, 
year={2006},
}

@article{rashedi2009,
author={E. Rashedi and  H. Nezamabadi-pour and S. Saryazdi}, 
title={GSA: A Gravitational Search Algorithm},
journal={Information Sciences}, 
volume={179}, 
number={13}, 
page={2232-2248}, 
month={June}, 
year={2009},
}

@inproceedings{oliveira2018, 
author={P. B. de Moura Oliveira and Josenalde Oliveira and José Boaventura Cunha}, 
title={Trends in Gravitational Search Algorithm}, 
booktitle={Distributed Computing and Artificial Intelligence, 14th International Conference, Advances in Intelligent Systems and Computing}, 
year={2018}, 
volume={620}, 
number={}, 
pages={270-277},
digitaldoi={10.1007/978-3-319-62410-5\_33}, 
ISSN={}, 
month={July},
}

@article{fgwcgsa,
author={I. H. Pamungkas and S. Pramana}, 
title={Improvement Method of Fuzzy Geographically Weighted Clustering using Gravitational Search Algorithm},
journal={Journal of Computer Science and Information}, 
volume={11}, 
number={1}, 
page={10-16}, 
month={June}, 
year={2019},
}

@article{randomfa,
author="Fister, Iztok
and Yang, Xin-She
and Brest, Janez
and Fister, Iztok",
title="On the Randomized Firefly Algorithm",
journal="Cuckoo Search and Firefly Algorithm: Theory and Applications",
year="2013",
publisher="Springer International Publishing",
address="Cham",
pages="27--48",
isbn="978-3-319-02141-6",
doi="10.1007/978-3-319-02141-6\_2",
url="https://doi.org/10.1007/978-3-319-02141-6\_2"
}

@article{intfa,
author={Fateen, Seif-Eddeen K and Bonilla-Petriciolet, Adrián},
title={Intelligent Firefly Algorithm for Global Optimization},
year={2013},
journal={Cuckoo Search and Firefly Algorithm: Theory and Applications},
volume={516},
pages={315-330},
}

@article{Karaboga2011,
  doi = {10.1016/j.asoc.2009.12.025},
  url = {https://doi.org/10.1016/j.asoc.2009.12.025},
  year = {2011},
  month = jan,
  publisher = {Elsevier {BV}},
  volume = {11},
  number = {1},
  pages = {652--657},
  author = {Dervis Karaboga and Celal Ozturk},
  title = {A novel clustering approach: Artificial Bee Colony ({ABC}) algorithm},
  journal = {Applied Soft Computing}
}

@INPROCEEDINGS{nurmala2015,
  author={N. {Nurmala} and A. {Purwarianti}},
  booktitle={2015 International Conference on Information Technology Systems and Innovation (ICITSI)}, 
  title={Improvement of Fuzzy Geographically Weighted Clustering-Ant Colony Optimization using context-based clustering}, 
  year={2015},
  volume={},
  number={},
  pages={1-6},
  doi={10.1109/ICITSI.2015.7437726}}

@article{Dhal2019,
  doi = {10.1007/s00521-019-04585-z},
  url = {https://doi.org/10.1007/s00521-019-04585-z},
  year = {2019},
  month = nov,
  publisher = {Springer Science and Business Media {LLC}},
  volume = {32},
  number = {8},
  pages = {3059--3077},
  author = {Krishna Gopal Dhal and Jorge G{\'{a}}lvez and Sanjoy Das},
  title = {Toward the modification of flower pollination algorithm in clustering-based image segmentation},
  journal = {Neural Computing and Applications}
}

@article{Nasution2020,
  doi = {10.1016/j.ijdrr.2020.101801},
  url = {https://doi.org/10.1016/j.ijdrr.2020.101801},
  year = {2020},
  month = dec,
  publisher = {Elsevier {BV}},
  volume = {51},
  pages = {101801},
  author = {Bahrul Ilmi Nasution and Robert Kurniawan and Tiodora Hadumaon Siagian and Ahmad Fudholi},
  title = {Revisiting social vulnerability analysis in Indonesia: An optimized spatial fuzzy clustering approach},
  journal = {International Journal of Disaster Risk Reduction}
}

@article{caraka2020cluster,
  title={Cluster Around Latent Variable for Vulnerability Towards Natural Hazards, Non-Natural Hazards, Social Hazards in West Papua},
  author={Caraka, Rezzy Eko and Lee, Youngjo and Chen, Rung Ching and Toharudin, Toni and Gio, Prana Ugiana and Kurniawan, Robert and Pardamean, Bens},
  journal={IEEE Access},
  volume={9},
  pages={1972--1986},
  year={2020},
  publisher={IEEE}
}

@article{setiawanrobert,
  title={Parameter Optimization of Support Vector Regression using Harris Hawks Optimization},
  author={ Setiawan, I Nyoman and Kurniawan, Robert and Yuniarto,Budi and  Caraka, Rezzy Eko and Pardamean, Bens},
  journal={Procedia Computer Science},
  volume={179},
  pages={17-24},
  year={2021},
  publisher={Elsevier}
}

@article{caraka2019pso,
  title={Prediction of status particulate matter 2.5 using state Markov chain stochastic process and HYBRID VAR-NN-PSO},
  author={Caraka, Rezzy Eko and Chen, Rung Ching and Toharudin, Toni and Pardamean, Bens and Yasin, Hasbi and Wu, Shih Hung},
  journal={IEEE Access},
  volume={7},
  pages={161654--161665},
  year={2019},
  publisher={IEEE}
}

@incollection{Bairathi2019,
  doi = {10.1007/978-3-030-16660-1\_81},
  url = {https://doi.org/10.1007/978-3-030-16660-1\_81},
  year = {2019},
  month = {apr},
  publisher = {Springer International Publishing},
  pages = {832--842},
  author = {Divya Bairathi and Dinesh Gopalani},
  title = {A Novel Swarm Intelligence Based Optimization Method: Harris' Hawk Optimization},
  booktitle = {Advances in Intelligent Systems and Computing}
}

@article{Heidari2019,
  doi = {10.1016/j.future.2019.02.028},
  url = {https://doi.org/10.1016/j.future.2019.02.028},
  year = {2019},
  month = aug,
  publisher = {Elsevier {BV}},
  volume = {97},
  pages = {849--872},
  author = {Ali Asghar Heidari and Seyedali Mirjalili and Hossam Faris and Ibrahim Aljarah and Majdi Mafarja and Huiling Chen},
  title = {Harris hawks optimization: Algorithm and applications},
  journal = {Future Generation Computer Systems}
}

@article{Kuo2018,
  doi = {10.1016/j.asoc.2018.02.039},
  url = {https://doi.org/10.1016/j.asoc.2018.02.039},
  year = {2018},
  month = jun,
  publisher = {Elsevier {BV}},
  volume = {67},
  pages = {299--308},
  author = {R.J. Kuo and T.C. Lin and F.E. Zulvia and C.Y. Tsai},
  title = {A hybrid metaheuristic and kernel intuitionistic fuzzy c-means algorithm for cluster analysis},
  journal = {Applied Soft Computing}
}

@inproceedings{Kennedy,
  doi = {10.1109/icnn.1995.488968},
  url = {https://doi.org/10.1109/icnn.1995.488968},
  publisher = {{IEEE}},
  year = {1995},
  pages = {1042-1948},
  author = {J. Kennedy and R. Eberhart},
  title = {Particle swarm optimization},
  booktitle = {Proceedings of {ICNN}{\textquotesingle}95 - International Conference on Neural Networks}
}

@article{Karaboga2007,
  doi = {10.1007/s10898-007-9149-x},
  url = {https://doi.org/10.1007/s10898-007-9149-x},
  year = {2007},
  month = apr,
  publisher = {Springer Science and Business Media {LLC}},
  volume = {39},
  number = {3},
  pages = {459--471},
  author = {Dervis Karaboga and Bahriye Basturk},
  title = {A powerful and efficient algorithm for numerical function optimization: artificial bee colony ({ABC}) algorithm},
  journal = {Journal of Global Optimization}
}

@incollection{Yang2009,
  doi = {10.1007/978-3-642-04944-6\_14},
  url = {https://doi.org/10.1007/978-3-642-04944-6\_14},
  year = {2009},
  publisher = {Springer Berlin Heidelberg},
  pages = {169--178},
  author = {Xin-She Yang},
  title = {Firefly Algorithms for Multimodal Optimization},
  booktitle = {Stochastic Algorithms: Foundations and Applications}
}

@incollection{Yang2012,
  doi = {10.1007/978-3-642-32894-7\_27},
  url = {https://doi.org/10.1007/978-3-642-32894-7\_27},
  year = {2012},
  publisher = {Springer Berlin Heidelberg},
  pages = {240--249},
  author = {Xin-She Yang},
  title = {Flower Pollination Algorithm for Global Optimization},
  booktitle = {Unconventional Computation and Natural Computation}
}

@article{Tsekouras2013,
  doi = {10.1016/j.fss.2012.10.004},
  url = {https://doi.org/10.1016/j.fss.2012.10.004},
  year = {2013},
  month = jun,
  publisher = {Elsevier {BV}},
  volume = {221},
  pages = {65--89},
  author = {George E. Tsekouras and John Tsimikas},
  title = {On training {RBF} neural networks using input{\textendash}output fuzzy clustering and particle swarm optimization},
  journal = {Fuzzy Sets and Systems}
}

@article{Rao2012,
  doi = {10.1080/0305215x.2011.652103},
  url = {https://doi.org/10.1080/0305215x.2011.652103},
  year = {2012},
  month = dec,
  publisher = {Informa {UK} Limited},
  volume = {44},
  number = {12},
  pages = {1447--1462},
  author = {R. V. Rao and V. J. Savsani and J. Balic},
  title = {Teaching{\textendash}learning-based optimization algorithm for unconstrained and constrained real-parameter optimization problems},
  journal = {Engineering Optimization}
}

@inproceedings{Bansal2011,
  doi = {10.1109/nabic.2011.6089659},
  url = {https://doi.org/10.1109/nabic.2011.6089659},
  year = {2011},
  month = oct,
  publisher = {{IEEE}},
  pages = {633-640},
  author = {J. C. Bansal and P. K. Singh and Mukesh Saraswat and Abhishek Verma and Shimpi Singh Jadon and Ajith Abraham},
  title = {Inertia Weight strategies in Particle Swarm Optimization},
  booktitle = {2011 Third World Congress on Nature and Biologically Inspired Computing}
}

@article{Grekousis2012,
  doi = {10.1016/j.apgeog.2011.11.004},
  url = {https://doi.org/10.1016/j.apgeog.2011.11.004},
  year = {2012},
  month = may,
  publisher = {Elsevier {BV}},
  volume = {34},
  pages = {125--136},
  author = {George Grekousis and Hatzichristos Thomas},
  title = {Comparison of two fuzzy algorithms in geodemographic segmentation analysis: The Fuzzy C-Means and Gustafson{\textendash}Kessel methods},
  journal = {Applied Geography}
}

@article{Spiezia2007,
  doi = {10.52324/001c.8300},
  url = {https://doi.org/10.52324/001c.8300},
  year = {2007},
  month = nov,
  publisher = {Southern Regional Science Association},
  author = {Vincenzo Spiezia and Stephan Weiler},
  title = {Understanding Regional Growth},
  journal = {Review of Regional Studies}
}

@phdthesis{Wijayanto2015a,
author = {Wijayanto, Arie Wahyu},
school = {Insitut Teknologi Bandung},
title = {{Improvement of fuzzy geo-demographic clustering using metaheuristic optimization on Indonesia population census}},
type = {Master's Thesis},
year = {2015}
}

@Manual{rdist,
    title = {rdist: Calculate Pairwise Distances},
    author = {Nello Blaser},
    year = {2020},
    note = {R package version 0.0.5},
    url = {https://CRAN.R-project.org/package=rdist},
}

@Article{ppclust,
    title = {Comparison of Internal Validity Indices for Fuzzy Clustering},
    author = {Zeynel Cebeci},
    journal = {Journal of Agricultural Informatics},
    year = {2019},
    number = {2},
    volume = {10},
    pages = {1-14},
    doi = {10.17700/jai.2019.10.2.537},
}

@article{Rao2012b,
author = {Rao, R. Venkata and Patel, Vivek},
doi = {10.5267/j.ijiec.2012.03.007},
issn = {19232926},
journal = {International Journal of Industrial Engineering Computations},
number = {4},
pages = {535--560},
title = {{An elitist teaching-learning-based optimization algorithm for solving complex constrained optimization problems}},
volume = {3},
year = {2012}
}

@INPROCEEDINGS{bhadoria2016,
  author={A. {Bhadoria} and M. {Singh} and M. {Gupta}},
  booktitle={2016 International Conference on Electrical, Electronics, and Optimization Techniques (ICEEOT)}, 
  title={An improved elitism based teaching-learning optimization algorithm}, 
  year={2016},
  volume={},
  number={},
  pages={3726-3730},
  doi={10.1109/ICEEOT.2016.7755407}}

@INPROCEEDINGS{li2017gsa,
  author={J. {Li} and N. {Dong}},
  booktitle={2017 13th International Conference on Computational Intelligence and Security (CIS)}, 
  title={Gravitational Search Algorithm with a New Technique}, 
  year={2017},
  volume={},
  number={},
  pages={516-519},
  doi={10.1109/CIS.2017.00120}}

@article{blum2003,
author = {Blum, Christian and Roli, Andrea},
title = {Metaheuristics in Combinatorial Optimization: Overview and Conceptual Comparison},
year = {2003},
issue_date = {September 2003},
publisher = {Association for Computing Machinery},
address = {New York, NY, USA},
volume = {35},
number = {3},
issn = {0360-0300},
url = {https://doi.org/10.1145/937503.937505},
doi = {10.1145/937503.937505},
journal = {ACM Comput. Surv.},
month = sep,
pages = {268–308},
numpages = {41}
}

@article{Fister2013,
archivePrefix = {arXiv},
arxivId = {1307.4186},
author = {Fister, Iztok and Yang, Xin-She and Fister, Iztok and Brest, Janez and Fister, Du{\v{s}}an},
eprint = {1307.4186},
journal = {Elektrotehniski Vestnik/Electrotechnical Review},
month = {jul},
number = {3},
pages = {116--122},
title = {{A Brief Review of Nature-Inspired Algorithms for Optimization}},
url = {http://arxiv.org/abs/1307.4186},
volume = {80},
year = {2013}
}

@Manual{naspaclustpkg,
title = {naspaclust: Nature-Inspired Spatial Clustering},
author = {Bahrul Ilmi Nasution and Robert Kurniawan and Rezzy Eko Caraka},
note = {R package version 0.1.1},
year = {2021},
url = {https://cran.r-project.org/package=naspaclust}
}

@incollection{Shahrudin2014,
  doi = {10.1007/978-3-642-41674-3\_105},
  url = {https://doi.org/10.1007/978-3-642-41674-3\_105},
  year = {2014},
  publisher = {Springer Berlin Heidelberg},
  pages = {735--741},
  author = {Muhammad Shahrizan Shahrudin and M. Mahmuddin},
  title = {Experiment on Modified Artificial Bee Colony for Better Global Optimisation},
  booktitle = {Lecture Notes in Electrical Engineering}
}

@article{NASUTIONfpa,
title = {Urban vulnerability to floods investigation in jakarta, Indonesia: A hybrid optimized fuzzy spatial clustering and news media analysis approach},
journal = {International Journal of Disaster Risk Reduction},
volume = {83},
pages = {103407},
year = {2022},
issn = {2212-4209},
doi = {https://doi.org/10.1016/j.ijdrr.2022.103407},
url = {https://www.sciencedirect.com/science/article/pii/S2212420922006264},
author = {Bahrul Ilmi Nasution and Fandi Muhammad Saputra and Robert Kurniawan and Ahmad Nurhadi Ridwan and Ahmad Fudholi and Bagus Sumargo}
}

@INPROCEEDINGS{nasutiontlbo,
  author={Nasution, Bahrul Ilmi and Siregar, Sri Indriyani},
  booktitle={2022 1st International Conference on Information System \& Information Technology (ICISIT)}, 
  title={Regional Economy Condition in Indonesia during COVID-19 Pandemic: An Analysis using Teaching Learning-Based Fuzzy Geodemographic Clustering}, 
  year={2022},
  volume={},
  number={},
  pages={37-42},
  doi={10.1109/ICISIT54091.2022.9873033}}

@article{KURNIAWANsvdata,
title = {Revisiting social vulnerability analysis in Indonesia data},
journal = {Data in Brief},
volume = {40},
pages = {107743},
year = {2022},
issn = {2352-3409},
doi = {https://doi.org/10.1016/j.dib.2021.107743},
url = {https://www.sciencedirect.com/science/article/pii/S2352340921010180},
author = {Robert Kurniawan and Bahrul Ilmi Nasution and Neli Agustina and Budi Yuniarto}
}

@article{Sumargo2024,
  title = {Green Infrastructure Vulnerability and Regional Poverty Reduction: New Sustainable Development Recommendations Based on a Spatial Clustering Approach},
  ISSN = {1540-7608},
  url = {http://dx.doi.org/10.1080/10875549.2024.2379769},
  DOI = {10.1080/10875549.2024.2379769},
  journal = {Journal of Poverty},
  publisher = {Informa UK Limited},
  author = {Sumargo,  Bagus and Kurniawan,  Robert and Nasution,  Bahrul Ilmi and Firmansyah,  Aldi and Laksono,  Bagaskoro Cahyo and Gio,  Prana Ugiana and Isnaeni,  Mohamad Andrian and Yusuf,  Mukhtar and Tarigan,  Vita Cita Emia},
  year = {2024},
  month = jul,
  pages = {1–24}
}

\end{document}